\documentclass[lettersize,journal]{IEEEtran}

\usepackage[fleqn]{amsmath}   
\usepackage{amssymb}
\usepackage{mathtools}
\usepackage{caption}
\usepackage{subcaption}
\usepackage[linesnumbered,ruled,vlined]{algorithm2e}
\usepackage[pdftex,pdfauthor={Marco Zangrandi, Stefano Arrigoni, Francesco Braghin},pdftitle={Control of a Hexapod Robot Considering Terrain Interaction}]{hyperref}

\begin{document}

\title{Control of a Hexapod Robot Considering Terrain Interaction}

\author{Marco Zangrandi, Stefano Arrigoni, Francesco Braghin \\
Politecnico di Milano, Mechanical Department, via G. La Masa, 1 - 20156 Milano Italy}
%
%

\maketitle

\begin{abstract}
Bio-inspired walking hexapod robots are a relatively young branch in robotics in both state of the art and applications. Despite their high degree of flexibility and adaptability derived by their redundant design, the research field that compliments their abilities is still very lacking. \\
In this paper will be proposed state-of-the-art hexapod robot specific control architecture that allows for full control over robot speed, body orientation and walk gait type to employ. \\
Furthermore terrain interaction will be deeply investigated, leading to the development of a terrain-adapting control algorithm that will allow the robot to react swiftly to terrain shape and asperities such as non-linearities and non-continuity within the workspace. It will be presented a dynamic model derived from the interpretation of the hexapod movement to be comparable to these of the base-platform PKM machines, and said model will be validated through Matlab SimMechanics\textsuperscript{TM} physics simulation. A feed-back control system able to recognise leg-terrain touch and react accordingly to assure movement stability will then be developed.
Finally results coming from an experimental campaign based of the PhantomX AX Metal Hexapod Mark II robotic platform by Trossen Robotics\textsuperscript{TM} is reported.
\end{abstract}

\begin{IEEEkeywords}
Hexapod Robot, Rough Terrain, Locomotion Control, Hexapod Dynamic Model
\end{IEEEkeywords}

\section{Introduction}

Bio-inspired robotics is a fairly new, still early in development branch of modern robotics. Control architectures coming from bio-inspired designs are able to perform complex tasks as to walk, swim, crawl, jump or even fly \cite{specialfeature} thanks to specific forms of movement, known as gaits, that can sometimes even surpass in efficiency conventional engineering.

For walking robots the first conventional-design competitor will always be the wheeled robot, widely spread in any real application from outer-planetary rover exploration and military drones to commercial, human transportation and materials handling vehicles. However walker robots can have an edge on rough, hostile terrains where wheels are limited by their ability of maintaining static friction with the ground at all times and cannot display the same manoeuvrability as their legs-equipped counterparts \cite{leggedrobotoverview}, which can instead both display greater stability and body control under a wider variety of terrains \cite{proprioceptive}. For this very reason most applications for legged robots come from moving through rubble and obstacles-filled environments such as for disaster rescue applications \cite{rescueapplications} or maintenance and repair applications in difficult to traverse mechanical environments \cite{Shero}.

The reason why hexapod solutions are so common in robotics as legged rovers is that six legs is the optimal number to get a fair number of statically stable walking gaits \cite{roboticwalknaturalterrain} (quadruped robots only get one, bipedal robots don't have any) to have good variety of movement options and speed while being able to cycle through them freely to better adapt to the user's needs. Also the fact that at all times three opposed legs at minimum are supporting the body assures that under normal operating conditions statically unstable poses are never reached.

\begin{figure}
\begin{center}
\includegraphics[width=0.4\textwidth]{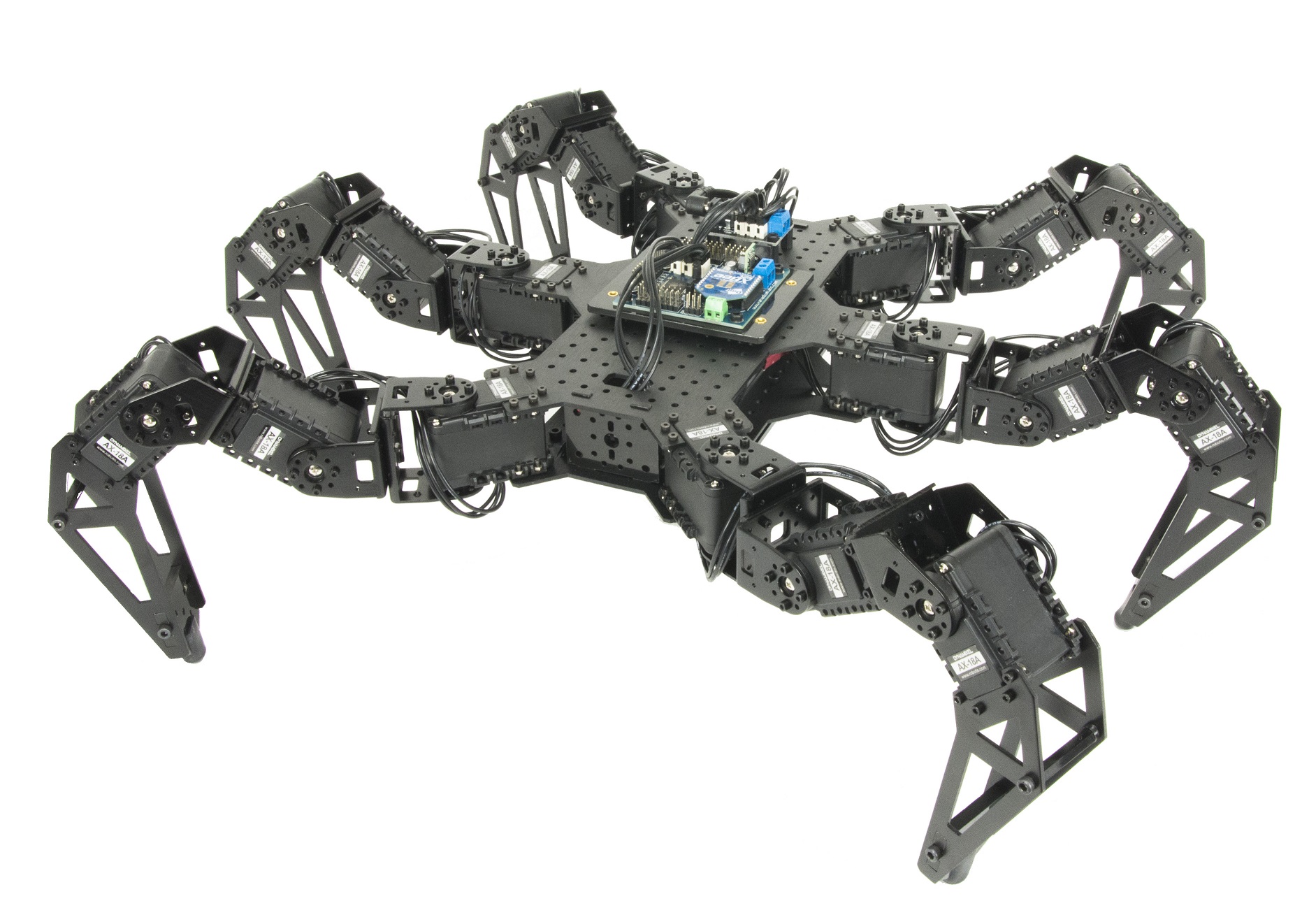}
\caption{PhantomX AX Metal Hexapod Mark II \cite{trossenhexapodweb} \label{phantomxpic}}
\end{center}
\end{figure}

The robotic platform selected for this research project is the PhantomX AX Metal Hexapod Mark II from Trossen Robotics \cite{trossenhexapodweb}, shown in Figure \ref{phantomxpic}. It mounts 18 Dynamixel AX-12A Smart Servomotors \cite{ax12aweb} that are able to provide feedback on position and load estimate as well as mounting large RAM banks to store information that allows them to be controlled with much more precision than of conventional servos.

Modelling a walker robot mechanics is a complex task \cite{robotmaniplegs, foot-terraininteractions, hexapodpredictive}. First, since there is no rigid connection from leg to the ground, either a RRR-joint is assumed and friction modelling is completely neglected \cite{humanoidwalkingrobot, virtualmodelbipedal} or a virtual spring-damper physics engine is written to simulate the interaction behaviour between legs and ground \cite{quadrupedrobotirregular, 3Dspringdamper}. In both cases though body push mechanics and relative displacements are never easy to describe in an easy formal way, and either a very complicated model is developed for a narrow range of applications \cite{impactforces} or a third-party physics engine is mounted in to aid the model-based control logic \cite{physicsengineterrain}.

Furthermore small, body relative micro-management displacements and poses are almost never considered fully, preferring to consider macro-movements to feed to trajectory finder algorithms and obstacle avoidance systems. Such is the case with \cite{hexapodpredictive} where a very complex model is developed to predict movement in a 3D environment and with \cite{octopus3} where trajectory definition and hexapod obstacle avoidance abilities are deeply investigated.

\subsection{Problem Assessment and Objectives}

The objective of the work proposed is the development of a a model that allows us to compensate macro terrain deformations and interfaces, to crawl through rough terrain sections and in general to perform these tasks automatically through minimal information about terrain environment and without user aid.

`NUKE' \cite{NUKEcodeonline} is the standard control software for any walker robot using AX-series and MX-series servos as main actuators that wills to implement a fast inverse kinematics engine. By asking the user to provide robot dimensions and to `capture' a `neutral' position of the robot the program generates custom locomotion software. \\
NUKE is able to generate a kinematic model that can automatically handle gait mechanism generation, robot body pose control and motion trajectory. Once the algorithm is compiled and uploaded to the control board, the robot can be commanded by user input and NUKE automatically handles walk motion cycle and servomotor control as to react to the user instructions in real time.

While NUKE allows for fast, real-time computation of an inverse kinematics engine, it suffers from the main limitation most multi-legged robot control software carry behind: While they are perfectly able to move the robot smoothly and adjust stride correctly through a gait engine, they are generally not able to interact with anything else than a flat, continuous, obstacle free terrain \cite{designissues}.

This paper takes as objective to extend the advantages of typical hexapod robots control software in presence of non-flat, non-continuous, irregular terrain. In particular the underlying objective is to develop a kinematic model able to account for terrain shape and location, and to further the analysis to see how robot physics interact with ground presence.

Finalizing a closed-loop control architecture is also essential to ensure robustness in movement especially under circumstances where gait stability is not guaranteed. Sloppy movement due to feed-forward control related errors and non-idealistic behaviours of the physical system can in fact destabilize the final robot pose. \\
In most application scenarios model-based control can be either greatly simplified or entirely skipped by accurately placing pressure sensors on its feet \cite{walkwithpressuresensors, forcesensorbasedwalking} and analysing pressure variation data to identify terrain contact \cite{complexorderdyn}. Furthermore it was shown in \cite{hexapodpositionfeedbackonly} how in some scenarios position feedback can be sufficient to accurately identify terrain presence and maintain horizontal positioning at all times. \\
However as our objectives aspire for greater controllability over the robot pose and reach autonomously movement stability under a number of tasks and situations a full, non reduced dynamic model will be employed. This means that the control software will be able to analyse a full-state feedback for robot torque and use it to provide the robot perception of its surroundings.

\section{Locomotion Control}

\begin{figure}
\begin{center}
\includegraphics[width=0.5\textwidth]{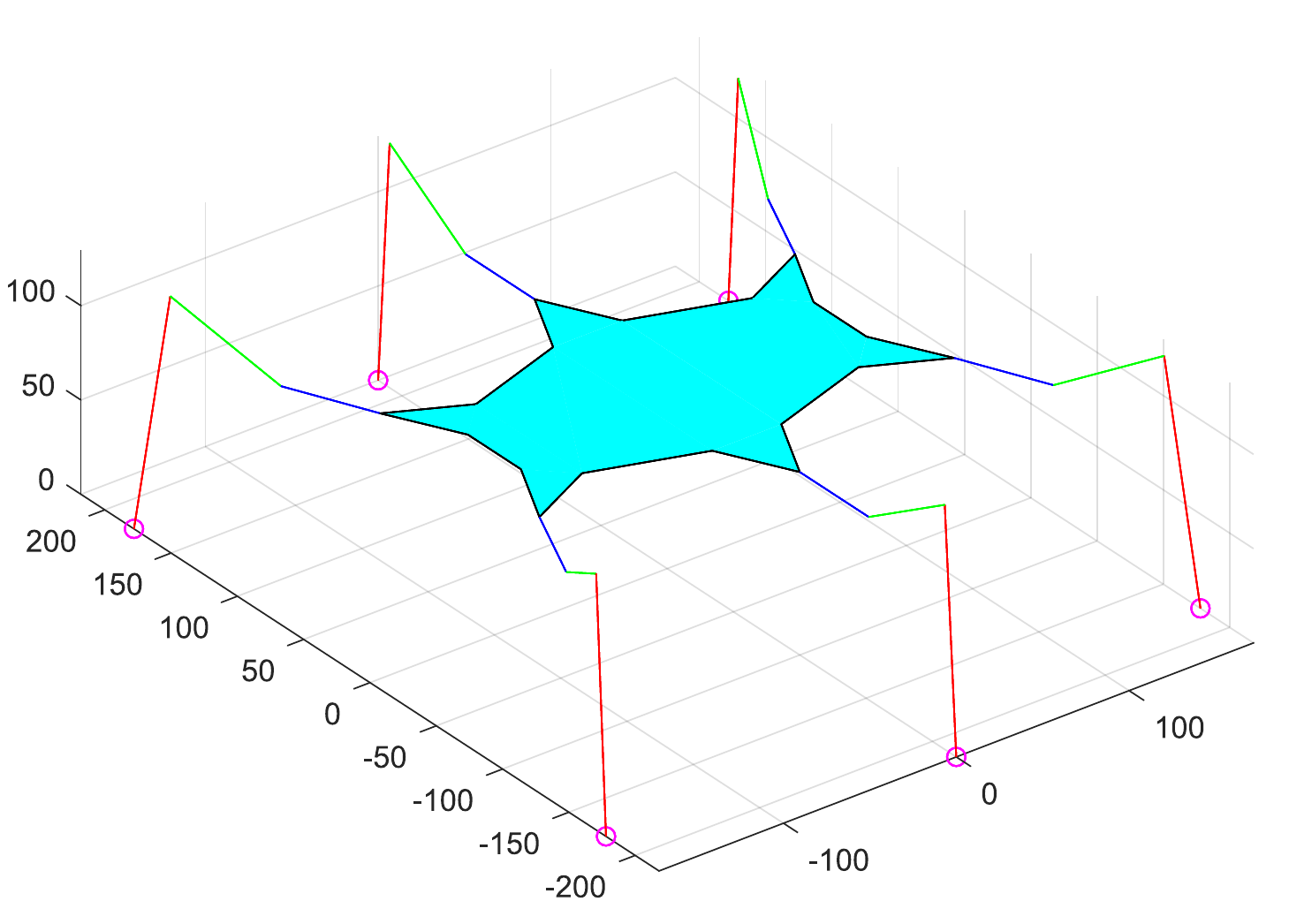}
\caption{Robot endpoints in neutral position\label{neutralposition}}
\end{center}
\end{figure}

Locomotion control of the hexapod crawler robot is mainly accomplished through the direct command of its legs endpoints. Legs endpoints are defined as the 3-D coordinate points located at the extremities of each leg as shown in Figure \ref{neutralposition}. \\
Since legs are 3-DOF systems that lead to the endpoints it is possible to attain full control of a leg by imposing its endpoint position and employing an inverse kinematics engine to calculate the associated servomotor angles.

The advantage coming from such control architecture is the possibility of collapsing the description of each leg configuration to a single point position in space, leading to a much simpler and clearer handling of the robot movement.

The position the robot assumes at the deployment state is called \emph{neutral position} and the associated endpoints coordinates are hardcoded into the robot controller software. This position represents a neutral state the kinematic engine will use as reference to build its walking gait.

\subsection{Endpoints Handling}

\begin{figure}
\begin{center}
\includegraphics[width=0.5\textwidth]{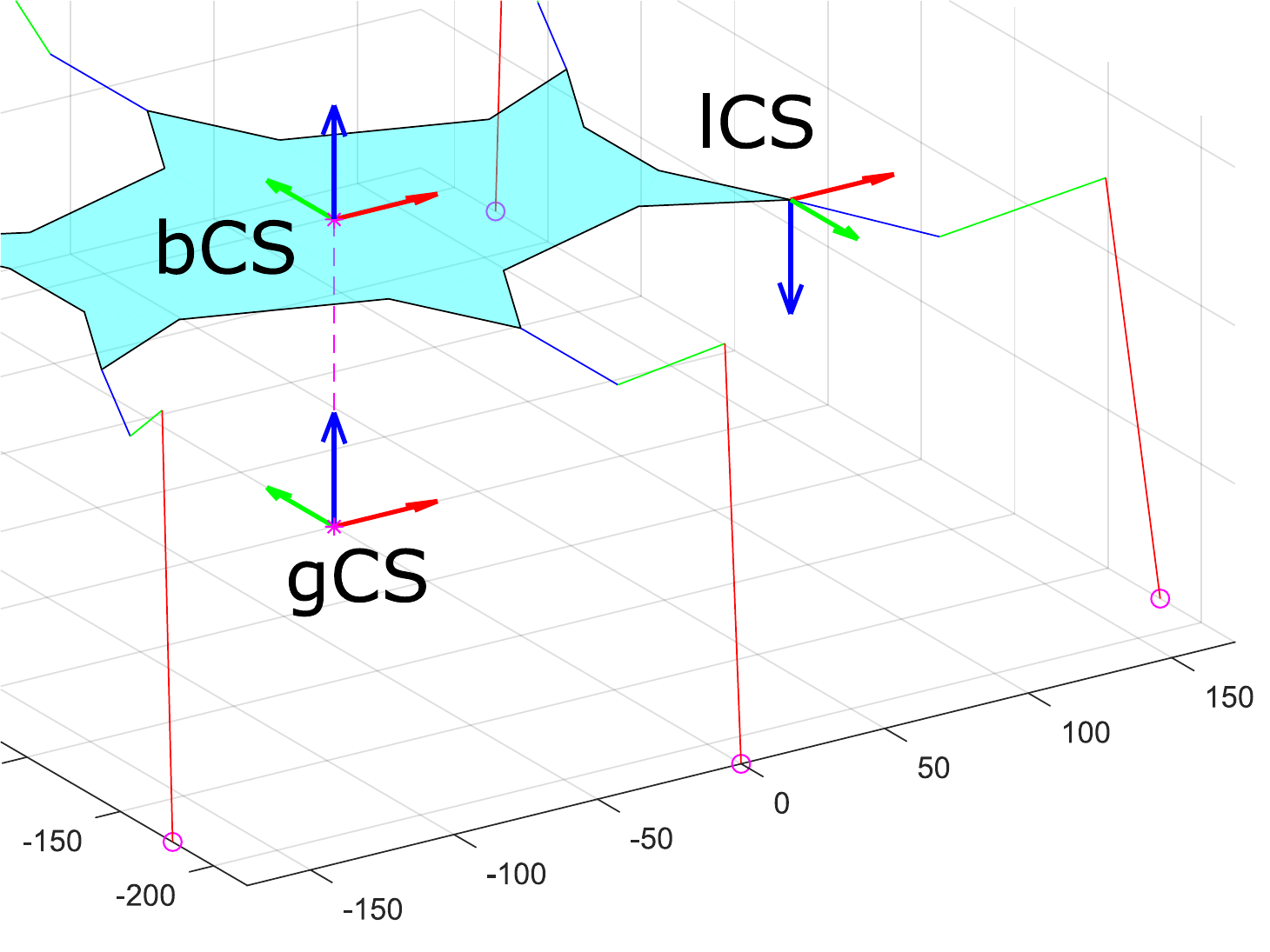}
\caption{The body, leg and global coordinate systems\label{allCS}}
\end{center}
\end{figure}

The coordinate systems employed in the kinematic analysis of the robot are presented in Figure \ref{allCS}. The \emph{global coordinate system} is fixed at terrain level at 0,0,0 coordinates and will be used to account for robot position with regard to the terrain and the environment. Therefore for the first iteration the transformation matrix for the global to body coordinate system $T_{globalbody,0}$ (alias: $T_{gb,0}$) will be as shown in \eqref{Tgb0}.

\begin{equation}
\label{Tgb0}
T_{gb,0} = \left[\begin{matrix}
1 & 0 & 0 & \mathrm{SP}_x \\
0 & 1 & 0 & \mathrm{SP}_y \\
0 & 0 & 1 & \mathrm{SP}_z \\
0 & 0 & 0 & 1
\end{matrix}\right]
\end{equation}

\noindent Where `SP' is the \emph{starting position}. Note that $\mathrm{SP}_z$ should be the initial robot height and must be consistent with the neutral position endpoints coordinates. \\
The \emph{body coordinate system} and the \emph{legs coordinate systems} instead move alongside the robot body and are used to both describe the robot orientation and solve the inverse kinematics for endpoint position and servomotor angles.

The transformation matrix that binds the leg c.s. and the body c.s. $T_{bodyleg}$ (as: $T_{bl}$)  is defined as \eqref{Tbodyleg}.

\begin{equation}
\label{Tbodyleg}
T_{bl} = \left[\begin{matrix}
1 & 0 & 0 & x_j \\
0 & -1 & 0 & y_j \\
0 & 0 & -1 & 0 \\
0 & 0 & 0 & 1
\end{matrix}\right]
\end{equation}

\noindent Where in \eqref{Tbodyleg} $x_j$ and $y_j$ are the joint position in body c.s., from which it is inferred that $T_{bl}$ is unique for each leg as each leg has a different joint position.

\begin{figure*}
	\begin{minipage}[c][\width]{0.4\textwidth}
	   \centering
	   \includegraphics[width=1\textwidth]{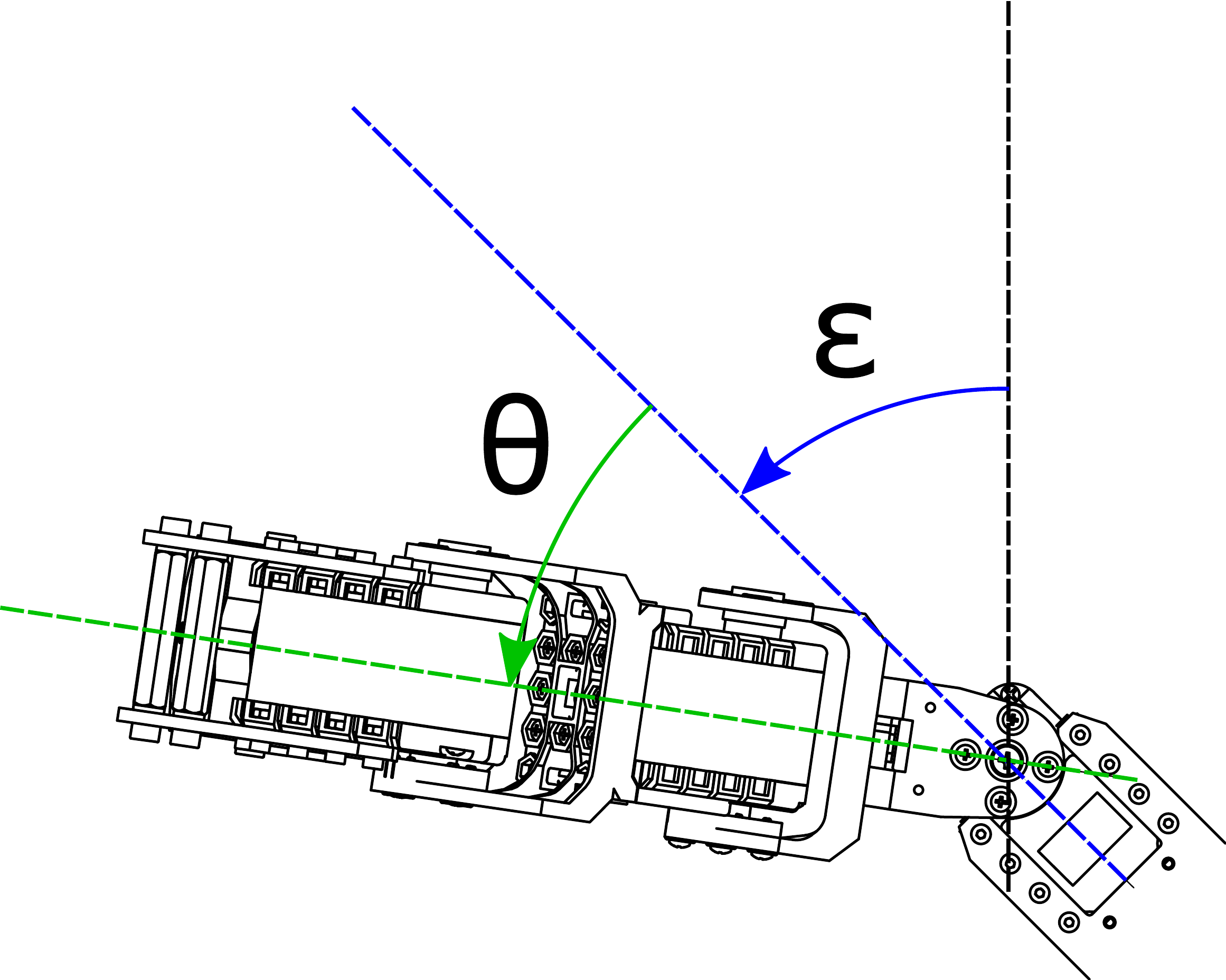}
	\end{minipage}
 \hfill 	
	\begin{minipage}[c][\width]{0.4\textwidth}
	   \centering
	   \includegraphics[width=1\textwidth]{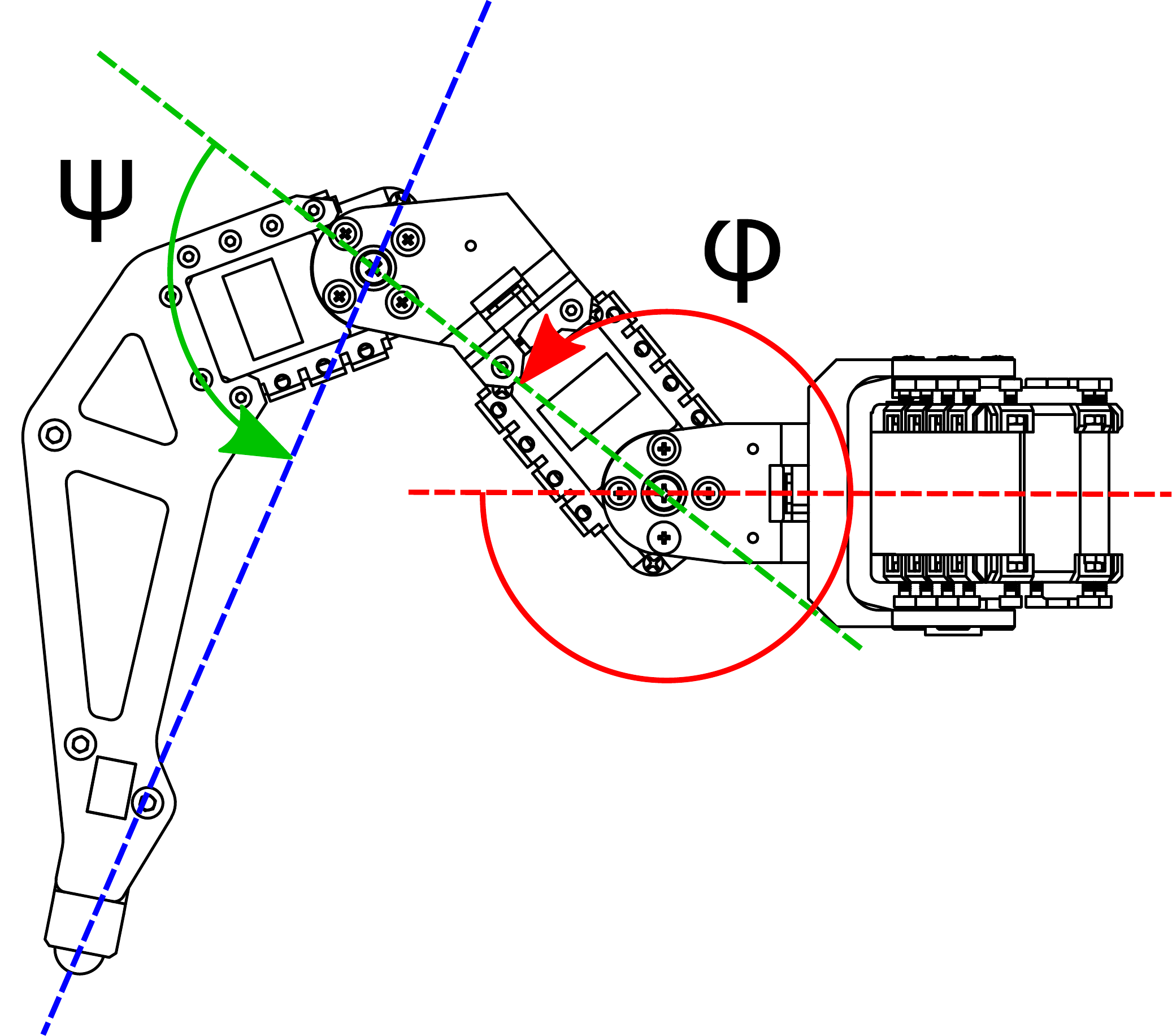}
	\end{minipage}
\caption{Definition of $\theta$, $\phi$ and $\psi$}
\label{thetaphipsi}
\end{figure*}

Once the desired endpoints position has been defined, servomotor angles can be assessed. The convention used in the kinematic analysis is shown in Figure \ref{thetaphipsi} where $\epsilon$ represents the servomotor orientation and $\theta$, $\phi$ and $\psi$ represent the coxa, femur and tibia servomotor angles respectively. \\
By expressing the desired endpoint position in leg coordinate system, eventually through \eqref{Tbodyleg}, servomotor angles can be calculated through \eqref{IKstart}--\eqref{IKend}.

\begin{align}
\label{IKstart}
&trueX = \sqrt{x^2 + y^2} - l_c \\
&im = \sqrt{trueX^2 + z^2} \\
&\theta = -\arctan\left(\frac{y}{x}\right) - \epsilon \\
&\phi = \frac{\pi}{2} - \arctan\left(\frac{trueX}{z}\right) - \arccos\left(\frac{l_f^2 + im^2 - l_t^2}{2\,im\,l_f}\right) \\
\label{IKend}
&\psi = \pi - \arccos\left(\frac{l_f^2 + l_t^2 - im^2}{2\,l_f\,l_t}\right)
\end{align}

\noindent Where in \eqref{IKstart}--\eqref{IKend} $(x,y,z)$ is the position of the desired leg endpoint in leg coordinate system and $l_c$, $l_f$ and $l_t$ are the lengths of the coxa, femur and tibia respectively.
Endpoints positions are defined in body coordinate system and therefore their significance is fully derivative to the position and orientation of the robot body. This means that every movement the robot body is instructed to go through can be immediately transposed to a relative displacement of the endpoints; the complete kinematic model can be built based on that assumption.

A gait engine is a subroutine of the locomotion algorithm that handles legs synchronization: In general any pedal locomotion divides each leg in either \emph{pushing} or \emph{swinging} state in fixed order in such a way to allow for a repetitive, continuous movement. The gait engine is responsible for assigning the \emph{swing} and \emph{push} roles as well as the direction and amount of space to cover for each iteration tick. The gait engine employed is not fundamental to describe the desired formulation and will not be discussed as already available in literature\footnote{The gait engine employed in this treatment is the one provided by NUKE.}. Therefore from the user input comprised of x-speed, y-speed and z-axis rotation speed the gait engine provides the movement data the robot is supposed to follow, the kinematic problem is to find the related endpoints displacement.

To solve the kinematic problem it is imperative to be able to describe the robot position and orientation at each iteration step by taking into consideration all movement data provided by the gait engine and by direct command of the user\footnote{The user is supposed to be able to bypass the gait engine instructions to apply direct control of the robot pose at each iteration step.}. \\
That is done by building the $T_{globalbody}$ transformation matrix at each $i^{th}$ iteration step (as: $T_{gb,i}$) as shown in \eqref{Tglobalbody}.

\begin{equation}
\label{Tglobalbody}
T_{gb,i} = T_{tr,i}\,Q_{ter,i}\,Mov_i\,RotZ_{b,i}\,RotY_i\,RotX_i
\end{equation}

\noindent Where in \eqref{Tglobalbody}:

\begin{itemize} 

\item $T_{tr,i}$ is the translational transformation matrix holding data about the movement instructions coming from the gait engine. It is built as shown in \eqref{Ttr} and reconstructs the robot position as if it were just under the influence of the gait engine alone.

\begin{equation}
\label{Ttr}
T_{tr,i} = T_{tr,i-1}\,Q_{tr,i} = T_{gb,0}\,\cdot\,Q_{tr,1}\,\cdot\,\ldots\,\cdot\,Q_{tr,i}
\end{equation}

\noindent In which $Q_{tr,i}$ is the transformation matrix representing the $i^{th}$ step movement due to the gait engine instructions. It is made of two contributions as shown in \eqref{Qtr}.

\begin{equation}
\label{Qtr}
Q_{tr,i} = \delta T_{mov,i}\,\delta RotZ_{g,i}
\end{equation}

\noindent In which $\delta T_{mov,i}$ is the component related to the translational movement and $\delta RotZ_{g,i}$ the one related to z-axis rotation. Those are defined in \eqref{deltaTmov} and \eqref{deltaRotZ} respectively.

\begin{equation}
\label{deltaTmov}
\delta T_{mov,i} = \left[\begin{matrix}
1 & 0 & 0 & \delta x_{gait,i} \\
0 & 1 & 0 & \delta y_{gait,i} \\
0 & 0 & 1 & 0 \\
0 & 0 & 0 & 1
\end{matrix}\right]
\end{equation}

\begin{equation}
\label{deltaRotZ}
\delta RotZ_{g,i} = \left[\begin{matrix}
\cos{\left(\delta rot_{z_{gait,i}}\right)} & -\sin{\left(\delta rot_{z_{gait,i}}\right)} & 0 & 0 \\
\sin{\left(\delta rot_{z_{gait,i}}\right)} & \cos{\left(\delta rot_{z_{gait,i}}\right)} & 0 & 0 \\
0 & 0 & 1 & 0 \\
0 & 0 & 0 & 1
\end{matrix}\right]
\end{equation}

\noindent Where in \eqref{deltaTmov}--\eqref{deltaRotZ} $\delta x_{gait,i}$, $\delta y_{gait,i}$ and $\delta rot_{z_{gait,i}}$ are the gait engine movement instructions for the $i^{th}$ step x-axis displacement, y-axis displacement and z-axis rotation respectively.

\item $Mov_i$ is the transformation matrix holding the user-imposed translational movement data of the robot body, defined as in \eqref{Mov}.

\begin{equation}
\label{Mov}
Mov_i = \left[\begin{matrix}
1 & 0 & 0 & bodyPos_{x_i} \\
0 & 1 & 0 & bodyPos_{y_i} \\
0 & 0 & 1 & bodyPos_{z_i} \\
0 & 0 & 0 & 1
\end{matrix}\right]
\end{equation}

\item $RotZ_{b,i}$, $RotY_i$ and $RotX_i$ are the transformation matrices holding user-imposed orientation of the robot body, being the z-axis rotational matrix, y-axis rotational matrix and x-axis rotational matrix respectively.

\item $Q_{ter,i}$ is the terrain-compensated reorientation matrix addressing additional body displacement due to terrain shape. It is dependant on the position of the robot body in the terrain environment and can either be given by user input or computed in real-time on basis of the terrain shape in the robot surroundings. Our objective is to assemble automatically this matrix by employing a terrain-tuning algorithm that takes as its only input the terrain elevation function $h(x,y)$ which can again either be given as user input (assuming perfect knowledge of the terrain shape) or constructed by an estimation architecture. $Q_{ter,i}$ is an identity matrix for completely flat terrains.

\end{itemize}

\noindent And therefore the full movement comprising of all contributions the robot body goes through at the $i^{th}$ iteration step is represented by the transformation matrix calculated as in \eqref{Qbody}.

\begin{equation}
\label{Qbody}
Q_{body,i} = T_{gb,i-1}^{-1}\,T_{gb,i}
\end{equation}

The \emph{pushing legs} endpoints are fixed to the terrain due to friction, therefore their global position should not change between iterations. This means that if the body moves as described by \eqref{Qbody}, then the relative position of the endpoints should change as \eqref{endpoints1}

\begin{equation}
\label{endpoints1}
\left[\begin{matrix} x_i \\ y_i \\ z_i \\ 1 \end{matrix}\right] =
Q_{body,i}^{-1}\,\left[\begin{matrix} x_{i-1} \\ y_{i-1} \\ z_{i-1} \\ 1 \end{matrix}\right]
\end{equation}

\noindent And so solving the kinematic problem for the \emph{pushing legs} endpoints. Note that since \eqref{endpoints1} is built recursively it needs starting values. That is the neutral position endpoints and this is the reason why their coordinates are hardcoded into the control software. \\
The \emph{swinging legs} endpoints positions are defined as \eqref{swingendp}.

\begin{equation}
\label{swingendp}
\left[\begin{matrix} x_i \\ y_i \\ z_i \\ 1 \end{matrix}\right]_{gCS} =
T_{tr,i}\,Q_{ter,i}\,RotZ_{g,i}\,T_{mov,i}\,
\left[\begin{matrix} x_{neu} \\ y_{neu} \\ z_{neu} \\ 1 \end{matrix}\right]_{bCS}
\end{equation}

\noindent Where $x_{neu}$, $y_{neu}$ and $z_{neu}$ are the endpoint coordinates of the neutral position. The reason why \eqref{swingendp} makes reference to the neutral position endpoint coordinates and addresses $RotZ_{g,i}$ and $T_{mov,i}$ instead of $\delta RotZ_{g,i}$ and $\delta T_{mov,i}$ is because of the core difference from \eqref{Tglobalbody} and \eqref{Qbody}: Instead of building the endpoints transformation from the previous endpoints position, the gait engine now needs to displace the leg in such a way that the swinging stride motion is obtained. This means that the movement data coming from $RotZ_{g,i}$ and $T_{mov,i}$ are no more about increments of movement but effective displacements from the neutral position. The gait engine should in fact return these values when assessing \emph{swinging state} legs endpoints. \\
To account for terrain presence and assure that the leg will always be over the terrain height while swinging, the correction shown in \eqref{swingcorrection} should be applied.

\begin{equation}
\label{swingcorrection}
z_{i,gCS} = z_{gait,i} + h\left(x_{i,gCS},y_{i,gCS}\right)
\end{equation}

\noindent Where $h(x,y)$ is the \emph{elevation function} of the terrain. \\
Once the global coordinates of the swinging legs endpoints are found, the body centred coordinates are found as in \eqref{swingbody}.

\begin{equation}
\label{swingbody}
\left[\begin{matrix} x_i \\ y_i \\ z_i \\ 1 \end{matrix}\right]_{bCS} = T_{gb,i}^{-1}\left[\begin{matrix} x_i \\ y_i \\ z_i \\ 1 \end{matrix}\right]_{gCS}
\end{equation}

\noindent Completing the kinematic problem assessment.

\subsection{Terrain Compensation Algorithm}

The objective of this section is to develop a way to orientate the body pose of the robot in such a way that while moving the robot freely on the ground, its body results `isolated' respect to the ground itself. \\
Assuming a perfect knowledge of terrain geometry in term of  angular and discontinuity interfaces position wouldn't be realistic and aligned with our objective of development of an adaptive algorithm. Our purpose is the design of an algorithm able to deal with any terrain just relying on the elevation function (measurable by on-board sensors mounted on the robot itself but not purpose of this work \cite{roughterrainmapping, negotiation}).

\begin{figure}
\begin{center}
\includegraphics[width=0.45\textwidth]{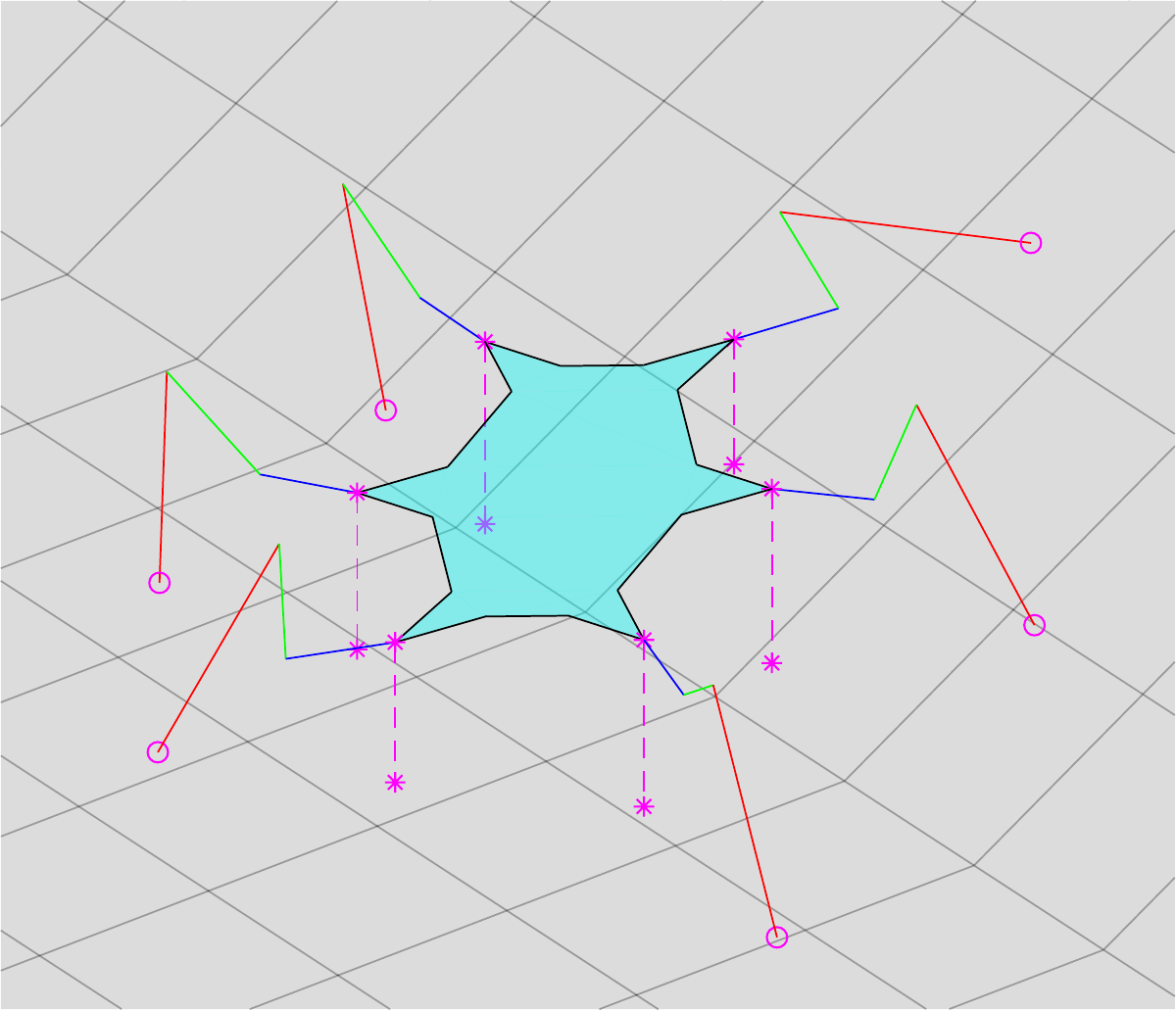}
\caption{Interest points positioned with robot shoulder joints\label{algopoints}}
\end{center}
\end{figure}

First of all it's important to unequivocally define what the `body isolation' condition is: This can be achieved by defining a set of points in the robot body and then consider the height of these points with respect to the ground as a way to evaluate the body relative position to the terrain. A possible approach is to ask for those points to maintain a distance to the ground as close as possible to the one defined by $\mathrm{SP}_z$ from \eqref{Tgb0}. If these points of interest are correctly chosen in order to represent the vertices of the robot body, the entire base should follow terrain profile and prevent unwanted situations like the ones discussed beforehand. \\
With the proposed algorithm the robot body will be able to position itself in such a way that it complies with the terrain geometry no matter the harshness. \\
In the algorithm presented six points are selected in the locations of the robot shoulders as shown in Figure \ref{algopoints}.

The main assumption of the compensation model is that the robot will be able to position itself in the optimal pose by moving from its horizontal pose defined by $T_{tr}$ with three degrees of freedom only: A first displacement in the z-direction followed by a rotation around its body c.s. y-axis and a final relative rotation around its body c.s. x-axis. In transformation matrices this is described by \eqref{Qterdef}.

\begin{equation}
\label{Qterdef}
Q_{ter,i} = D_{ter,i}\,RotY_{ter,i}\,RotX_{ter,i}
\end{equation}

\noindent Where:

\begin{itemize}

\item $D_{ter,i}$ is the z-axis rigid translation matrix defined as in \eqref{Dterdef}.

\begin{equation}
\label{Dterdef}
Disp_{ter,i} = \left[\begin{matrix}
1 & 0 & 0 & 0 \\
0 & 1 & 0 & 0 \\
0 & 0 & 1 & \delta z_{ter,i} \\
0 & 0 & 0 & 1
\end{matrix}\right]
\end{equation}

\item $RotY_{ter,i}$ and $RotX_{ter,i}$ are the rotation matrices defined as in \eqref{rot1} and \eqref{rot2}.

\begin{align}
\label{rot1}
RotY_{ter,i} &= Rot\left(\alpha_{ter,i},Y\right) \\
\label{rot2}
RotX_{ter,i} &= Rot\left(\beta_{ter,i},X\right)
\end{align}

\noindent Where in \eqref{rot1}--\eqref{rot2} $\alpha_{ter,i}$ is the angular rotation around the body c.s. y-axis and $\beta_{ter,i}$ is the angular rotation around the body c.s. x-axis.

\end{itemize}

From the global coordinate system the joint positions after the terrain reorientation are \eqref{shoulderrotationsgCS}.

\begin{equation}
\label{shoulderrotationsgCS}
\left[\begin{matrix} x_{j_{ter,i}} \\ y_{j_{ter,i}} \\ z_{j_{ter,i}} \\ 1 \end{matrix}\right]_{gCS}
= T_{tr,i}\,Disp_{ter_i}\,RotY_{ter_i}\,RotX_{ter_i}
\left[\begin{matrix} x_j \\ y_j \\ z_j \\ 1 \end{matrix}\right]_{bCS}
\end{equation}

\noindent Since we want our points to have relative heights as close to $\mathrm{SP}_z$ as possible we can build an algorithm that minimizes the total relative height quadratic error. It is accomplished accounting for each point, shown in \eqref{costfunction} with reference to \eqref{algosubstitutions}.

\begin{align}
\label{costfunction}
f &= \sum \left(z_{j_{ter}} - Z_{terrain} - \mathrm{SP}_z\right)^2 = \notag \\
&= \sum \epsilon_j^2 \left(a,b,c,d,e\right) = \notag \\
&= \sum (a - x_j\,b + z_j\,c\,d + y_j\,d\,e + \notag \\ 
&\qquad\qquad\qquad\qquad - h\left(x_{j_{ter}},y_{j_{ter}}) - \mathrm{SP}_z\right)^2 \\
\label{algosubstitutions}
&\left\{\begin{matrix}
a = \delta z_{ter,i} \\
b = \sin{\alpha_{ter,i}} \\
c = \cos{\beta_{ter,i}} \\
d = \cos{\alpha_{ter,i}} \\
e = \sin{\beta_{ter,i}}
\end{matrix}\right.
\end{align}

\noindent The problem is a multi-variable optimization problem with equality constraints coming from trigonometric functions consistency, solvable by writing the Newton-Euler equations with Lagrangian multipliers \cite{optimalcontrol}. The formulation obtained and reported in \eqref{costfunctionandlmult} comes from adjoining equality constraints \eqref{algosubstitutions} to cost function in \eqref{costfunction}.

comes from the adjoining to the cost function \eqref{costfunction} the equality constraints to get the Lagrangian function \eqref{costfunctionandlmult}.

\begin{equation}
\label{costfunctionandlmult}
L = f + \left[\begin{matrix} \lambda_1 & \lambda_2\end{matrix}\right]\left[\begin{matrix} c^2 + e^2 - 1 \\ b^2 + d^2 - 1\end{matrix}\right]
\end{equation}

\noindent Necessary conditions hold for the minimum \cite{optimalcontrol} as: $\nabla L = 0$. \\
The gradient of the Lagrangian function is calculated with the partial derivatives shown in \eqref{partialderivativesfirst}--\eqref{partialderivativeslast}.

\begin{align}
\label{partialderivativesfirst}
\frac{\partial L}{\partial a} &= \sum 2\epsilon_j \left(a,b,c,d,e\right) \\
\frac{\partial L}{\partial b} &= \sum 2\epsilon_j \left(a,b,c,d,e\right)(-x_j) + 2\lambda_2\,b \\
\frac{\partial L}{\partial c} &= \sum 2\epsilon_j \left(a,b,c,d,e\right)(z_j\,d) + 2\lambda_1\,c \\
\frac{\partial L}{\partial d} &= \sum 2\epsilon_j \left(a,b,c,d,e\right)(z_j\,c + y_j\,e) + 2\lambda_2\,d \\
\frac{\partial L}{\partial e} &= \sum 2\epsilon_j \left(a,b,c,d,e\right)(y_j\,d) + 2\lambda_1\,e \\
\frac{\partial L}{\partial \lambda_1} &= c^2 + e^2 - 1 \\
\label{partialderivativeslast}
\frac{\partial L}{\partial \lambda_2} &= b^2 + d^2 - 1 
\end{align}

\noindent In order to find the solution to this M-V-O-P numerical methods need to be employed. In this situation being the cost function quadratic it is possible to employ effectively the steepest descent algorithm \cite{Zunino} to iterate and find the solution very quickly and with a relatively low computational cost.


\noindent Having already defined the gradient of the Lagrangian, the steepest descent algorithm is implemented as in \eqref{algonextiter}.

\begin{equation}
\label{algonextiter}
X_{it+1} = X_{it} - \nabla L(X_{it}) \cdot Weight
\end{equation}

\begin{figure}
\includegraphics[width=0.45\textwidth]{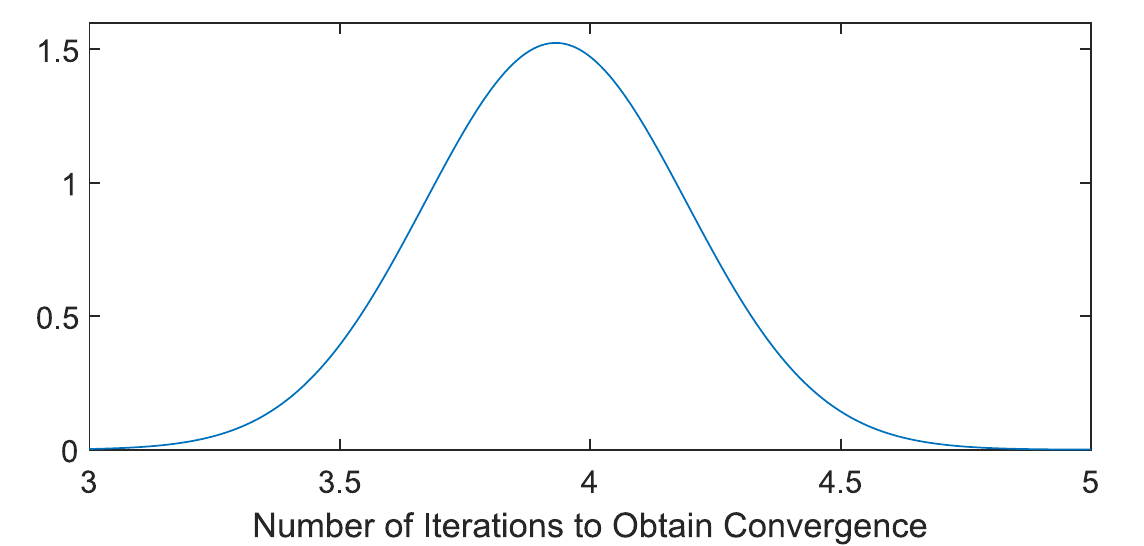}
\caption{Gaussian distribution of iterations needed to get to convergence with $Weight=1e-5$ and $tol=0.01$ for a typical walk task in the terrain $h(x,y) = 50\,mm. \times \left(\sin{\left(\frac{x}{100\,mm.}\right)} +  \cos{\left(\frac{y}{100\,mm.}\right)}\right)$\label{gaussian}}
\end{figure}

\noindent And stopping condition being $||X_{it} - X_{it-1}|| \leq tol$ . In \eqref{algonextiter} the $Weight$ parameter represents the adjustment index to the increment to the next-iteration solution. Generally a $Weight$ of $1e-5$ is good enough to get convergent solutions in a few iterations in most general applications. \\
This method assures good performances in most terrain situations, only experiencing non-convergence problems in extreme scenarios. As shown in Figure \ref{gaussian}, with reasonable weight ($Weight$) and tolerance ($tol$) the solution is generally found very quickly.

\begin{figure}
\includegraphics[width=0.45\textwidth]{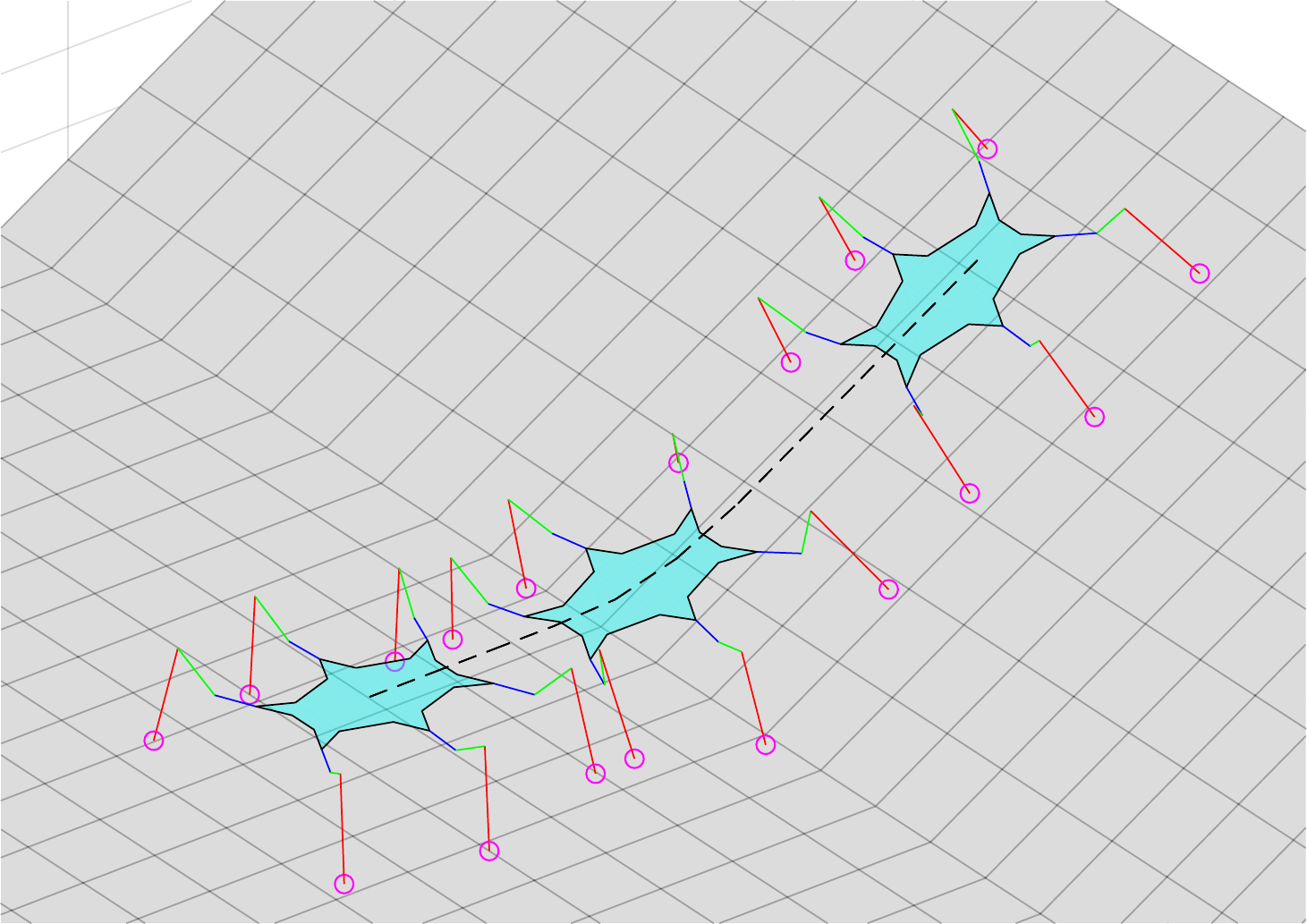}
\caption{Robot continuous adaptation to interface and ramp}
\label{terraincompfig}
\end{figure}

\begin{figure}[h!]
\includegraphics[width=0.45\textwidth]{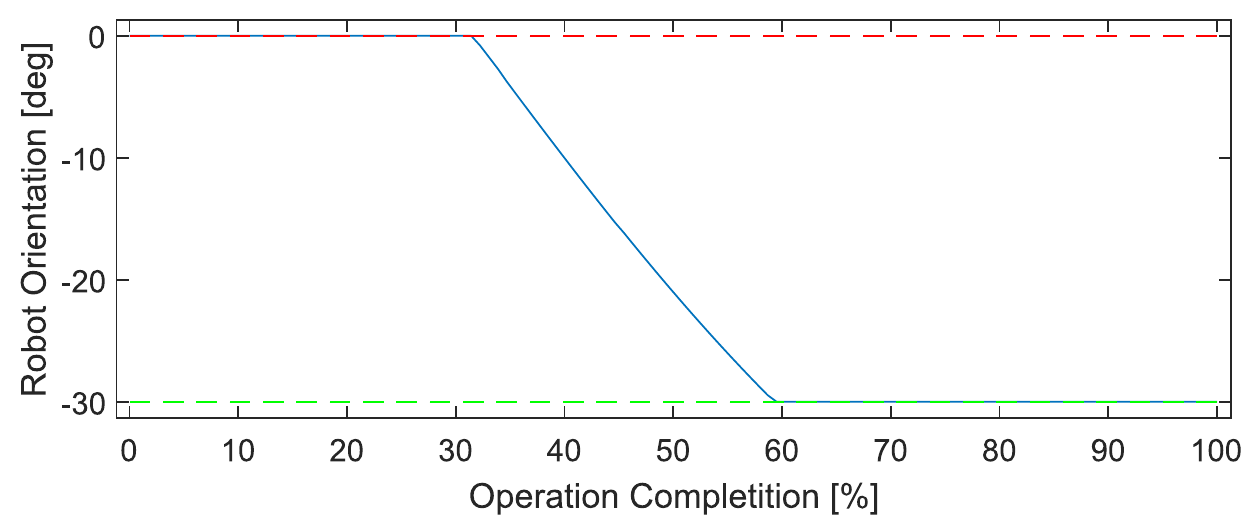}
\caption{Angle variation of robot inclination during ramp interface. Note how it is a continuous distribution until it matches the ramp's angulation.}
\label{terraincompgraph}
\end{figure}

Since the problem was defined in general terms, the algorithm is capable of running even in situations where rough interfaces are present, moving the robot in such a way to allow for a smooth transition between angular interfaces and navigating over non-continuous terrains. In the scenario of a steep ramp given a full horizontal speed input, the robot is able to smoothly go from being completely horizontal to adapting to the terrain inclination as shown in Figures \ref{terraincompfig} and \ref{terraincompgraph}.

\newpage

\noindent Note that since the algorithm runs on the elevation function only, realistic terrains defined through conventional discrete models \cite{digitalterrainmodels} can be used without needing particular adjustments.
\section{Torque Estimation}

\subsection{Dynamic Model}

Smart servos generally employed in physical hexapods are able to provide feed-back in both position and speed, however these signals are almost completely insensitive to terrain interaction effects; in particular they are blind to sliding and instability issues. Moreover they might not being affected by trips or grip losses. Even in overstepping cases, the robot will simply fall down in a rigid manner while reporting to be working perfectly. \\
For this reason to intercept terrain contact another signal must be taken into account. Torque is a value that is highly dependant on which legs are supporting the body or not, as well as reacting swiftly to terrain interaction. Servomotor torque can therefore become our way to make the robot inspect its surroundings and by compensating unstable-poses scenarios by repositioning the legs correctly we can assure robust movement through the entire control operation. \\
However to correctly interpret torque values coming from servos we need to develop a full dynamic model of the robot movement, as presented in the following section.

The degrees of freedom of the robot body are defined as in \eqref{Xpdef}.

\begin{equation}
\label{Xpdef}
\mathcal{X}_b = \left[\begin{matrix} x_b \\ y_b \\ z_b \\ rot_x \\ rot_y \\ rot_z \end{matrix}\right]
\end{equation}

\noindent Where the DOF follow the order set in \eqref{Tglobalbody}. \\
The degrees of freedom of a single leg are defined as the state vector shown in \eqref{Xaidef} with consistency to the previous definitions of $\theta$, $\phi$ and $\psi$.

\begin{equation}
\label{Xaidef}
\mathcal{X}_{a,l} = \left[\begin{matrix} \theta_i \\ \phi_i \\ \psi_i \end{matrix}\right]
\end{equation}

\noindent The state vector comprising of all legs degrees of freedom is defined as $\mathcal{X}_a$ shown in \eqref{Xadef}.

\begin{equation}
\label{Xadef}
\mathcal{X}_a = \left[\begin{matrix} \mathcal{X}_{a,1} \\ \vdots \\ \mathcal{X}_{a,6} \end{matrix}\right]
\end{equation}

The main problem coming with the modelling of the hexapod robot and legged locomotion in general is that to use lightweight dynamic algorithms like Newton-Euler equations it's an absolute necessity to have one and only one grounded joint at all times \cite{rigidbodyalgo} \cite{Legnani} \cite{Briot}. Legged locomotion usually does not fall under these requirements and multiple ground constraints need to be addressed which add great complexity to the model. \\
To avoid these kind of problems it is possible to take an alternate route for modelling: instead of considering the robot positionally constrained at the ground at the pushing legs endpoints we consider the robot as not having any constraints at all, and adjoining the kinematic constraints to the Lagrangian dynamic equations. \\
Since the ground contact points act as hinges for the robot, the kinematic constraints will be the nullity of these points' linear velocities.

\begin{figure}
\begin{center}
\includegraphics[width=0.4\textwidth]{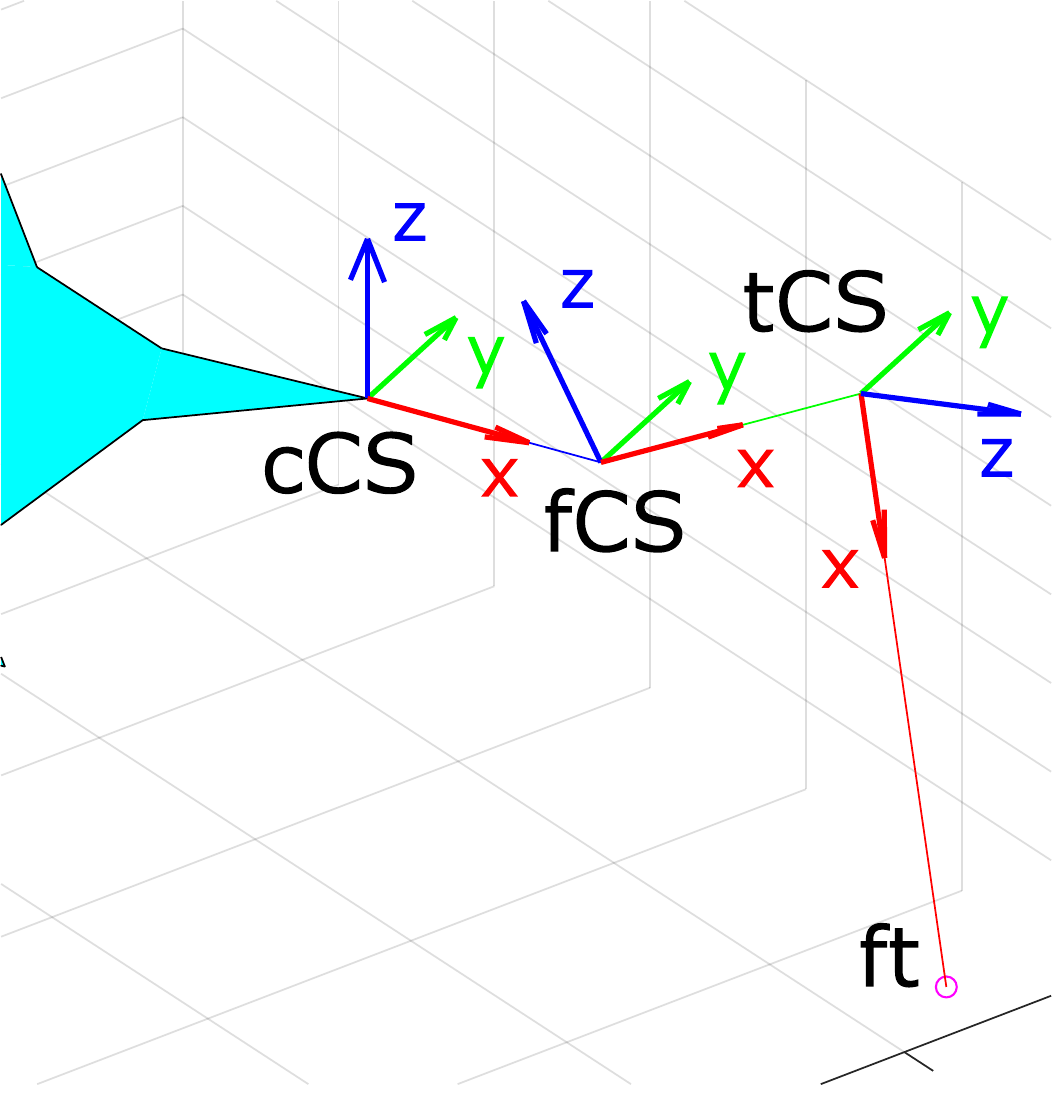}
\caption{Coxa, femur and tibia coordinate systems\label{cftCS}}
\end{center}
\end{figure}

The coxa, femur and tibia coordinate systems with reference to Figure \ref{cftCS} are identified with the matrices \eqref{Tbodycoxa}, \eqref{Tcoxafemur} and \eqref{Tfemurtibia}. Those are used to find the relative positions of either the coxa, femur and tibia joints or the feet endpoint, e.g. $r_{f-c}$ being the position of the femur joint in coxa coordinate system.

\begin{align}
\label{Tbodycoxa}
&T_{bodycoxa} = \left[\begin{matrix}
\cos{(\epsilon + \theta)} & -\sin{(\epsilon + \theta)} & 0 & x_j \\
\sin{(\epsilon + \theta)} & \cos{(\epsilon + \theta)} & 0 & y_j \\
0 & 0 & 1 & 0 \\
0 & 0 & 0 & 1 \end{matrix}\right] \\
\label{Tcoxafemur}
&T_{coxafemur} = \left[\begin{matrix}
\cos{\phi} & 0 & \sin{\phi} & l_c \\
0 & 1 & 0 & 0 \\
-\sin{\phi} & 0 & \cos{\phi} & 0 \\
0 & 0 & 0 & 1 \end{matrix}\right] \\
\label{Tfemurtibia}
&T_{femurtibia} = \left[\begin{matrix}
\cos{\psi} & 0 & \sin{\psi} & l_f \\
0 & 1 & 0 & 0 \\
-\sin{\psi} & 0 & \cos{\psi} & 0 \\
0 & 0 & 0 & 1 \end{matrix}\right]
\end{align}

By defining as $T_{trbody}$ in \eqref{Ttrbody} the transformation matrix that describes the effects of the body DOF on its pose, the axes of rotation of the coxa, femur and tibia joints can be found under a fixed coordinate system as shown in \eqref{Afirst}--\eqref{Alast}.

\begin{equation}
\label{Ttrbody}
T_{trbody} = Mov\,RotZ\,RotY\,RotX
\end{equation}

\begin{align}
\label{Afirst}
&A_{c} = T_{trbody}\,T_{bodycoxa}\,A_z \\
&A_{f} = T_{trbody}\,T_{bodyfemur}\,A_y \\
\label{Alast}
&A_{t} = T_{trbody}\,T_{bodytibia}\,A_y
\end{align}

\begin{figure}
\begin{center}
\includegraphics[width=0.4\textwidth]{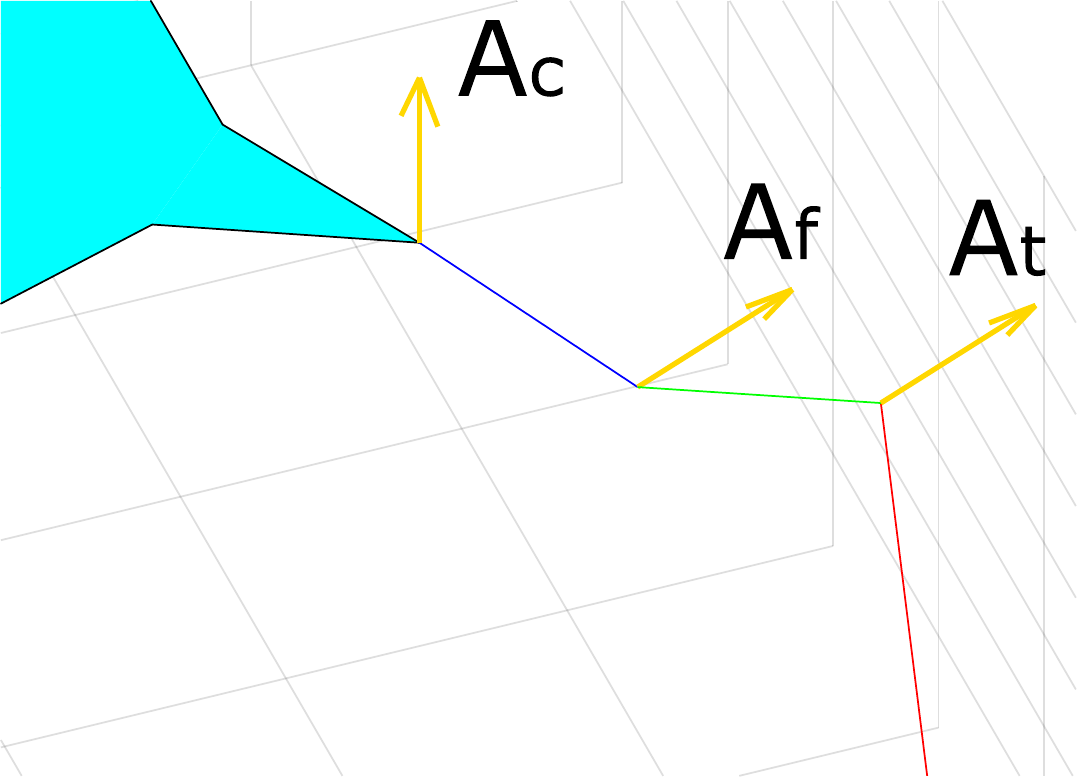}
\caption{Axes of rotations of actuated joints\label{actuatoraxes}}
\end{center}
\end{figure}

\noindent Where $A_z$ is the 0,0,1,0 axis and $A_y$ the 0,1,0,0 axis.

From this definition we can define the kinematic constrain for pushing legs grounded endpoints as \eqref{kinematicconstraint}, where it's trivial the derivation of \eqref{dotXadef} which shows the relation between the velocities of the actuators and the robot body DOF.

\begin{align}
\label{kinematicconstraint}
&A_{c,l}\,\dot{\mathcal{X}}_b + B_{c,l}\,\dot{\mathcal{X}}_{a,l} = 0 \\
\label{dotXadef}
&\dot{\mathcal{X}}_{a,l} = -B_{c,l}^{-1}\,A_{c,l}\,\dot{\mathcal{X}}_b
\end{align}

\noindent The $B_{c,l}$ and $B_{c,l}$ Jacobian constraint matrices definition is reported in \eqref{Bc} and \eqref{Ac}.

\begin{align}
\label{Bc}
&\!\!\!\!\!\!\!\!\!\!\!\!\!\!\!B_{c,l} = \left[\begin{array}{c|c|c} -\hat{r}_{ft-c}\,A_{c} & -\hat{r}_{ft-f}\,A_{f} & -\hat{r}_{ft-t}\,A_{t} \end{array}\right] \\
\label{Ac}
&\!\!\!\!\!\!\!\!\!\!\!\!\!\!\!A_{c,l} = \left[\begin{array}{c|c} \left(RotZ\,RotY\,RotX\right) & -\hat{r}_{f-b}\left(RotZ\,RotY\,RotX\right) \end{array}\right] \Psi
\end{align}

\noindent In which $\Psi$ is the matrix that transforms the time derivatives of the robot body DOF into the robot body twist\footnote{`twist' being the name given by \cite{Briot} of the kinematic screw and $\hat{r}$ is the skew matrix representation of the $r$ position vector. The kinematic screw is the velocity vector field of dimension $6 \times 1$ composed of the linear velocities and the angular velocities respectively.}.

A single constraint equation for all degrees of freedom of the robot requires to apply the definitions provided by \eqref{AcTOTdef} and \eqref{BcTOTdef} which result in the expression \eqref{kinematicconstraintTOT}.

\begin{align}
\label{AcTOTdef}
&A_c = \left[\begin{matrix} ip_1\,A_{c,1} \\ \vdots \\ ip_6\,A_{c,6} \end{matrix}\right] \\
\label{BcTOTdef}
&B_c = \left[\begin{matrix} ip_1\,B_{c,1} & & \\ & \ddots & \\ & & ip_6\,B_{c,6} \end{matrix}\right]
\end{align}

\begin{equation}
\label{kinematicconstraintTOT}
A_c\,\dot{\mathcal{X}}_b + B_c\,\dot{\mathcal{X}}_a = 0
\end{equation}

\noindent Where $ip_i$ is a boolean value that accounts for whether the $i^{th}$ leg is in pushing, constrained state (and therefore its endpoint is grounded) or in swinging state.

Since the kinematic constraint equation \eqref{kinematicconstraintTOT} was written with reference to both $\mathcal{X}_a$ and $\mathcal{X}_b$, the Lagrange equations will consider the full state vector as in \eqref{fullstate}.

\begin{equation}
\label{fullstate}
\mathcal{X} = \left[\begin{array}{c}\mathcal{X}_a \\ \hline \mathcal{X}_b \end{array}\right]
\end{equation}

\noindent By defining the kinematic constraint equation, the dynamic model is expressed as of \eqref{dyneq}.

\begin{equation}
\label{dyneq}
\tau = \tau_a + \left(-A_{c}^{+}\,B_{c}\right)^T \, \tau_b
\end{equation}

\noindent Where $\tau_a$ are the contributions coming from the real actuated joints $\mathcal{X}_a$ and $\tau_b$ are the ones coming from the body DOF $\mathcal{X}_b$. $A_{c}^{+}$ is the left pseudoinverse of the $A_{c}$ matrix.

By defining the inertia matrix as in \eqref{massmatrix} and the potential energies as \eqref{potentialen} \cite{Briot}, then the two contributions can be calculated as of \eqref{taua}--\eqref{taub}.

\begin{align}
\begin{split}
\label{massmatrix}
&\!\!\!\!\!\!\!\!\!\!\!\!M(\mathcal{X}) =\sum_{l=1}^6 \left( J_{c,l}^T\,M_{c}\,J_{c,l} + J_{f,l}^T\,M_{f}\,J_{f,l} + J_{t,l}^T\,M_{t}\,J_{t,l} \right) + \\
&\!\!\!\!\!\!\!\!\!\!\!\!\qquad\; + J_b^T\,M_{b}\,J_b = \left[\begin{array}{c} M_a(\mathcal{X}) \\ \hline M_b(\mathcal{X}) \end{array}\right]
\end{split} \\
\begin{split}
\label{potentialen}
&\!\!\!\!\!\!\!\!\!\!\!\!U = -\left[\begin{matrix} g^T & 0 \end{matrix}\right]\sum_{l=1}^6 (m_{c}\,T_{trbody}\,T_{bodycoxa,l}\,r_{g-c} + \\
&\!\!\!\!\!\!\!\!\!\!\!\!\qquad+ m_{f}\,T_{trbody}\,T_{bodyfemur,l}\,r_{g-f} + \\
&\!\!\!\!\!\!\!\!\!\!\!\!\qquad+m_{t}\,T_{trbody}\,T_{bodytibia,l}\,r_{g-t}) + \\
&\!\!\!\!\!\!\!\!\!\!\!\!\qquad -\left[\begin{matrix} g^T & 0 \end{matrix}\right]m_{b}\,T_{trbody}\,r_{g-b}
\end{split}
\end{align}


\begin{align}
\label{taua}
\begin{split}
&\!\!\!\!\!\!\!\!\!\!\!\!\tau_a = M_a \ddot{\mathcal{X}} + \left( \left(\mathcal{X}^T \otimes I_{18}\right) \frac{\partial M_a}{\partial \mathcal{X}} + \right.\\
&\!\!\!\!\!\!\!\!\!\!\!\!\qquad\qquad \left. - \frac{1}{2} \left(I_{18} \otimes \dot{\mathcal{X}}^T\right) \frac{\partial M}{\partial \mathcal{X}_{a}} \right) \dot{\mathcal{X}} + \left(\frac{\partial U}{\partial \mathcal{X}_a}\right)^T
\end{split} \\
\label{taub}
\begin{split}
&\!\!\!\!\!\!\!\!\!\!\!\!\tau_b = M_b \ddot{\mathcal{X}} + \left( \left(\mathcal{X}^T \otimes I_{6}\right) \frac{\partial M_b}{\partial \mathcal{X}} + \right. \\
&\!\!\!\!\!\!\!\!\!\!\!\!\qquad\qquad \left. - \frac{1}{2} \left(I_{6} \otimes \dot{\mathcal{X}}^T\right) \frac{\partial M}{\partial \mathcal{X}_b} \right) \dot{\mathcal{X}} + \left(\frac{\partial U}{\partial \mathcal{X}_b}\right)^T
\end{split}
\end{align}

\noindent In which $\otimes$ is the Kronecker product following notation by \cite{Taghirad}. Note that Since $\mathcal{X}_a$ is measurable and $\mathcal{X}_b$ can be estimated through $Q_{body,i}$ calculated through \eqref{Qbody}, the full state vector knowledge is assumed at all times.

\subsection{Dynamic Model Validation}

\begin{figure}[b]
\begin{center}
\includegraphics[width=0.4\textwidth]{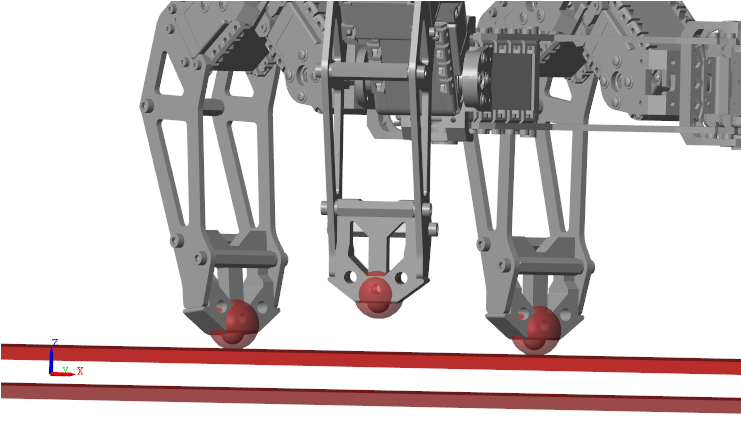}
\caption{Graphical rendition of the plane-spheres interaction through the Simscape Multibody Contact Forces Library\label{contactforceslib}}
\end{center}
\end{figure}

\begin{figure}
\includegraphics[width=0.5\textwidth]{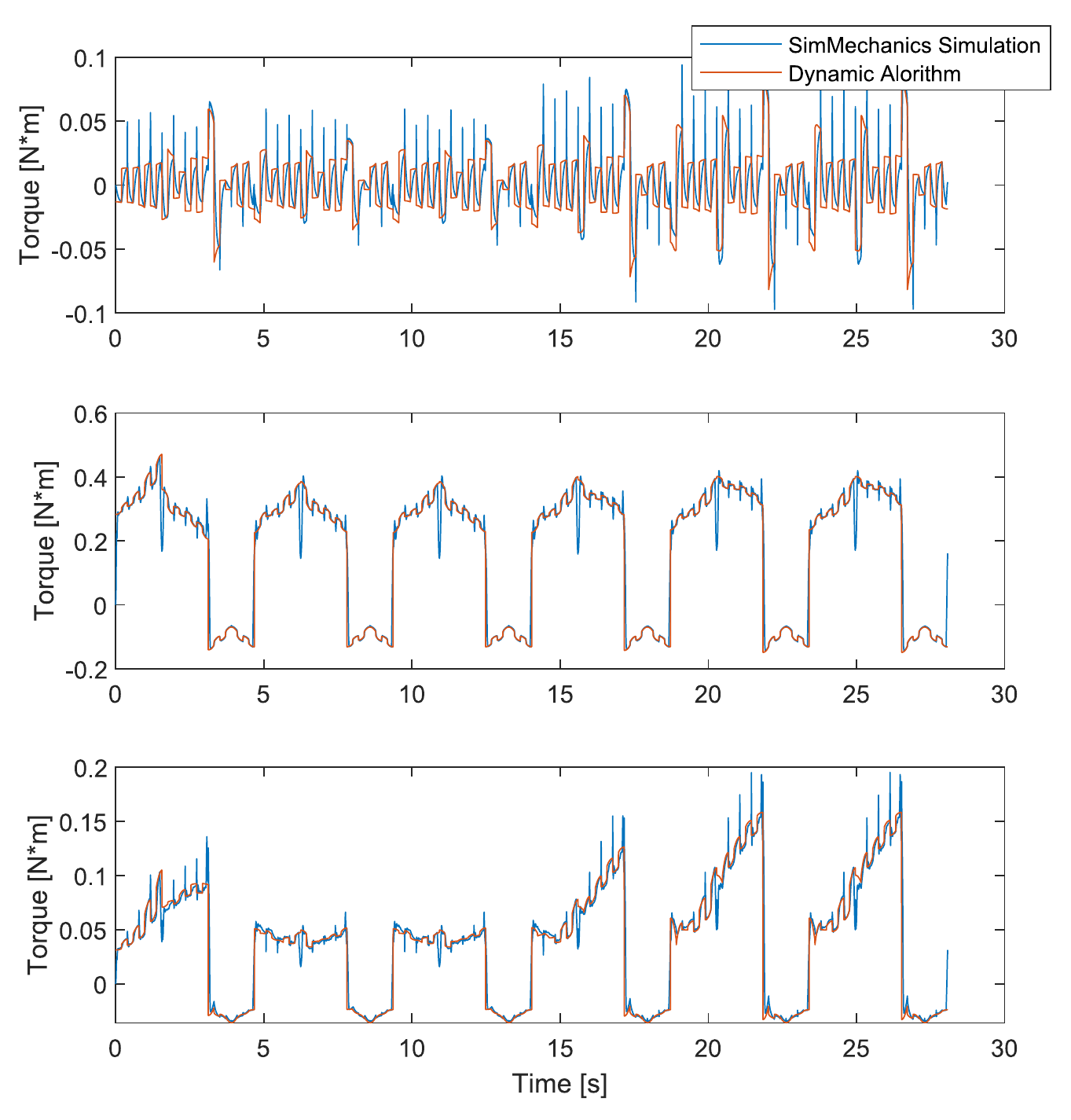}
\caption{Comparisons of dynamic results of a simple walk task from SimMechanics\textsuperscript{TM} simulation and the dynamic model. Graphs are referring to right-front leg's \emph{coxa}, \emph{femur} and \emph{tibia} actuators respectively.}
\label{dyncomp}
\end{figure}

The dynamic model was developed through various simplifying assumptions and so its validity is subordinate to the fact that those assumptions and simplifications are small enough to be neglected in a real application perspective. Following other works that deal with crawler robot physics \cite{HexapodSimMechanics} the dynamic model validation will be carried out through direct comparison of results with a Matlab SimMechanics\textsuperscript{TM} simulation. \\
The Matlab SimMechanics\textsuperscript{TM} simulation employs a physics engine and uses a full 3D model of the hexapod robot, as well as fully simulating friction and terrain physics through the Simscape Multibody Contact Forces Library \cite{simscapecontactforces} (foot-terrain interaction system shown in Figure \ref{contactforceslib}).


We can clearly see that from comparisons shown in Figure \ref{dyncomp} that the dynamic model closely matches the result coming from the SimMechanics\textsuperscript{TM} physics engine: In fact while there are differences in torque spikes coming from the simulated friction model physics, the overall behaviour of the torque perceived by the SimMechanics\textsuperscript{TM} physics engine closely matches the one generated by our dynamic algorithm. This not only means that our algorithm is indeed accurate enough to provide good results, but the assumptions we made while developing the dynamic model, such as the way we modelled grounded endpoints through kinematic constraints, the PKM interpretation of the hexapod movement and the simplification of the hexapod feet providing single-point contact between leg and terrain are small enough not to cause mismodelling errors in the final calculations.

\section{Closed-Loop Control}

A closed loop control approach is essential to obtain stable, robust movement throughout the entirety of the movement tasks the robot will be subjected to. A completely feed-forward approach may lead after few steps to positioning errors related mainly to robot-interface mismatching between reality and expected values, which would translate to grip losses and other unwanted behaviours to avoid in presence of uneven terrain.


The two major situations to avoid are the case where the feed-forward locomotion control system expects the terrain to be higher, and therefore leaves the leg hanging, and the case where it expects it to be lower, and therefore pushes the body back in an overstepping action destabilizing the entire robot pose. These situations are bound to happen due to either terrain mismodelling or desync between expected behaviour from simulation environment and actual robot motion, which is mainly due to the increasing drift between real and expected position of the robot as it accumulates positioning errors due to all non-ideal behaviours and performances coming from its hardware and components. Note how both situation happen in the leg-lowering part of the swing phase \emph{only}, meaning a feed-back control algorithm acting on real ground touching only needs to be called during that movement section.

%

\begin{figure}[t]
\begin{center}
\includegraphics[width=0.5\textwidth]{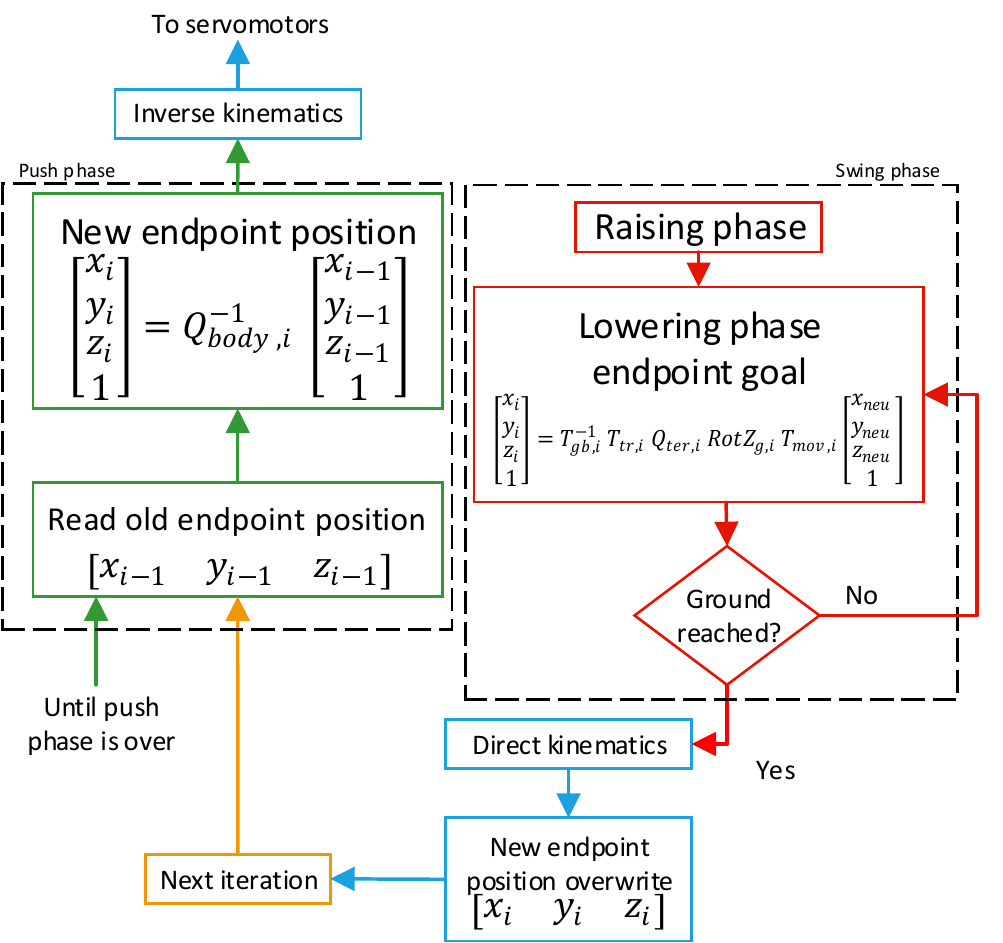}
\caption{Scheme of the feed-back control algorithm interaction with \emph{push} and \emph{swing} phases\label{FBgraph}}
\end{center}
\end{figure}

The way the feed-back control system works is by comparing the expected value of the actuators torques coming from the dynamic model and the actual sensed torques coming from the sensors system. \\
By reading the actuator position and speed and estimating body movement from $Q_{body,i}$ it is in fact possible at each iteration to build the estimated $\mathcal{X}_a$ and $\mathcal{X}_b$ state vectors, to which immediately calculate the $\tau_a$ and $tau_b$ components as defined in \eqref{taua}--\eqref{taub} being them the only actual dependencies. \\
Once these two values are stored in memory, the lowering legs' sudden terrain touch can be simulated in the dynamic model by simply interacting with the $ip_i$ values defined earlier in \eqref{AcTOTdef}--\eqref{BcTOTdef}. In fact the $ip_i$ boolean represents the grounded state of the $i^{th}$ leg's endpoint; therefore by temporarily setting it to 1 and building the $A_c$ and $B_c$ constraint matrices it is possible to get the hypothetical terrain-reached torques for all actuators through \eqref{dyneq}. This is an easy, relatively CPU-light way to get full references for the legs terrain-touch situations. \\
If there is a sensible correspondence between the expected torques and the sensed ones then the terrain is considered reached for that particular leg and its movements are stopped. This accounts for the overstepping part of the problem, while it may still be possible for legs to reach its desired end positions while no terrain having been sensed. \\
In order to solve the `leg-hanging' problem, it is only necessary to lower the expected altitude of the desired leg endpoint position\footnote{In the employed algorithm, the gait altitude instruction is lowered by 1 cm. each time.} $z_{gait,i}$ and recompute the needed servomotor angles. This can be obtained lowering the legs until terrain is touched and the previous condition is met, effectively assuring that legs reach ground and stop right on before moving to the \emph{push} phase. \\
Finally, since the servomotors stopping condition is changed to sensed terrain touch, lowering legs endpoints will very probably differ to the ones expected by the locomotion system instructions. Therefore the new positions need to be updated employing the forward kinematics tools such as the $T_{bodycoxa}$, $T_{bodyfemur}$ and $T_{bodytibia}$ transformation matrices derived through \eqref{Tbodycoxa}--\eqref{Tfemurtibia}. \\
A scheme summarizing the legs handling system is shown in Figure \ref{FBgraph}. An implementation example of the closed loop control solution is presented in Algorithm \ref{closedloopalgo}.

\begin{algorithm}[t]
\SetAlgoLined
 tau\_a, tau\_b = dynamic\_model(X\_a, X\_b, DynValues, TrMatrices)\;
 torques = read\_motor\_torques()\;
 \While{lowering\_legs \textbf{is not} empty}{
  \For{leg \textbf{in} lowering\_legs}{
   ip[leg] = 1\;
   A\_c, B\_c = constr\_matr(TrMatrices, ip)\;
   tau = tau\_a + (-pinv(A\_c)*B\_c).T * tau\_b\;
   \uIf{corrispondence(tau, torques) == 1}{
    stop\_motors(leg)\;
    theta, phi, psi = read\_motor\_angles(leg)\;
    endp\_pos[leg] = FK(theta, phi, psi, leg)\;
    \textbf{del} leg \textbf{in} lowering\_legs
   }
    \uElseIf{is\_motors\_stop(leg)}{
    endp\_pos[leg].z = endp\_pos[leg].z - 10\;
    }
    \uElse{
    \textbf{pass}\;
    }
   }
  }
 \caption{lowering legs control algorithm pseudo-code\label{closedloopalgo}}
\end{algorithm}
\section{Experimental Results}


The developed motion control architecture has been experimentally verified through the accomplishment of a series of testing scenarios. The experimental tests conducted have the aim to stress two important aspects of the architecture presented and highlight the improvements respect to a stock controller as well as to analyse the effect of feed-forward locomotion control and feed-back stabilization contribution. In particular the two scenarios reported are:

\begin{itemize}

\item \emph{An obstacle-filled terrain}: Where the robot is fed with a flat terrain data, and it needs to adapt to the presence of software `invisible' terrain. The task will be considered completed if the robot is able to remain stable due to overstepping into obstacles and tripping, completing the traversal correctly.

\item \emph{A Ramp terrain situation}: Where the robot is fed with data of the ramp terrain while the body is asked to remain horizontal throughout the full movement. The task will be considered completed if the robot is able to handle the angular interface, to climb the ramp effectively, and maintain the body horizontal at all times. Difficulty comes from the fact that even a small deviation in trajectory direction, which is highly possible since in the robotic platform no positional feedbacks are enforced, can cause significant errors between real and expected position of the terrain.

\end{itemize}

\begin{figure}[t]
	\begin{minipage}[c]{0.4\textwidth}
	   \centering
	   \includegraphics[width=\textwidth]{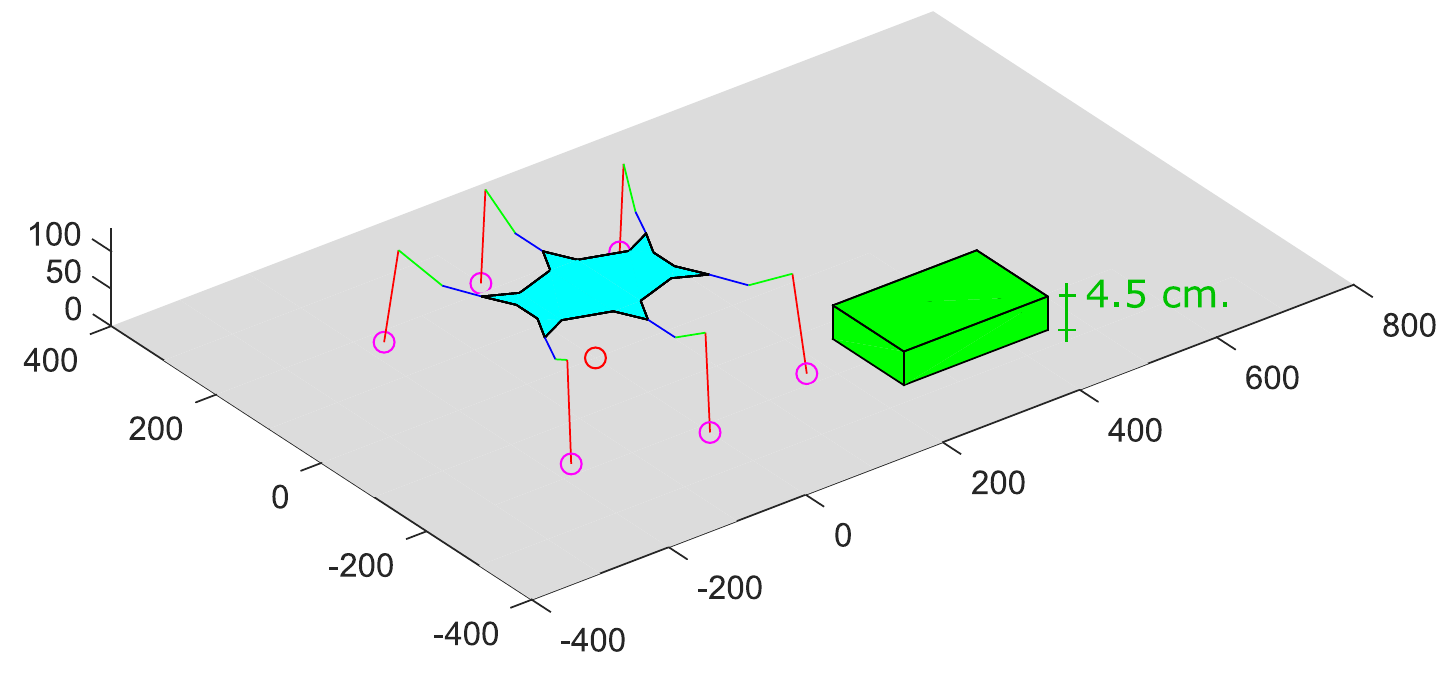}
	\end{minipage}
 \vskip\baselineskip	
	\begin{minipage}[c]{0.4\textwidth}
	   \centering
	   \includegraphics[width=\textwidth]{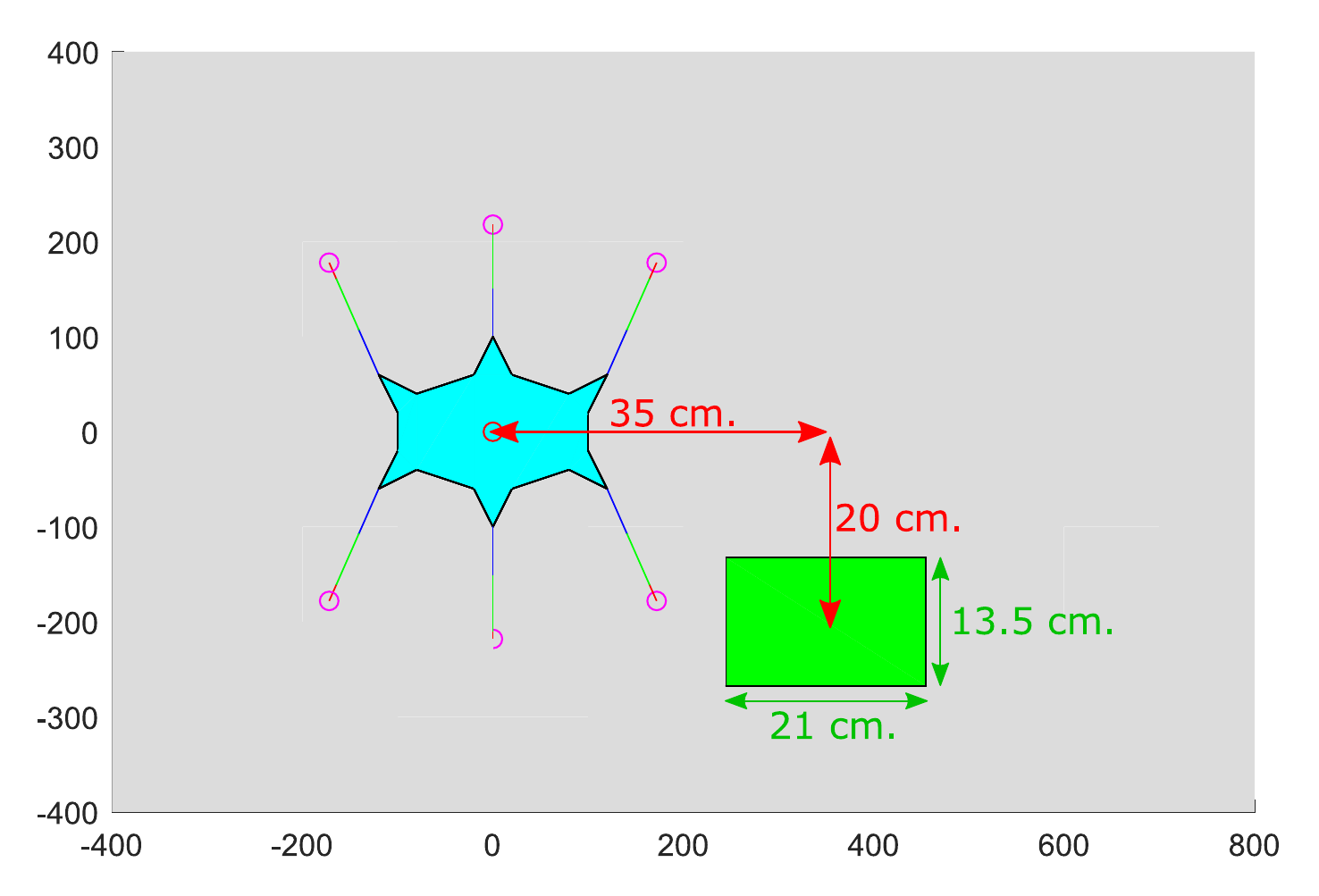}
	\end{minipage}
\caption{The obstacle course setup dimension and position}
\label{popperbookscheme}
\end{figure}


\subsection{Obstacle Recognition}

Feed-back control algorithm adds to the robot walking motion robustness as it allows to even sense and react to unexpected obstacles in the workspace. This means that when facing a mismodelled obstacle, like a section of the terrain wrongly modelled as flat and being instead of higher altitude, the feed-back logic is expected to compensate feed-forward error and and to update the gait in such a way to maintain movement stability.

In order to reproduce this condition a solid object was placed within the walk path of the robot. The obstacle and robot setup schematics are shown in Figure \ref{popperbookscheme}.

\begin{figure}[t]
	\begin{subfigure}{0.5\textwidth}
		\centering
		\begin{minipage}[c]{0.49\textwidth}
			\centering
			\includegraphics[width=\textwidth]{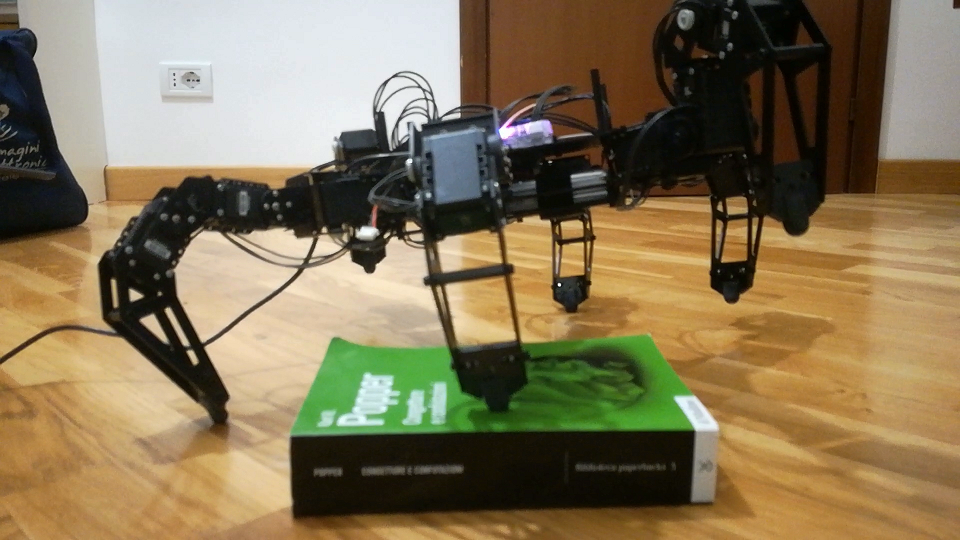}
		\end{minipage}
	\hfill
		\begin{minipage}[c]{0.49\textwidth}  
			\centering 
			\includegraphics[width=\textwidth]{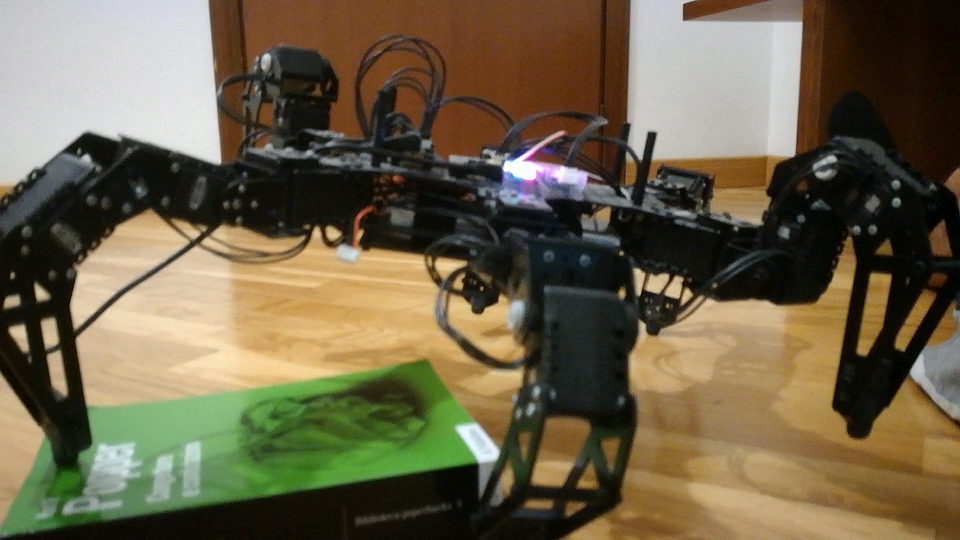}
		\end{minipage}
		\caption{Stock NUKE controller instabilities}
		\label{obstacletaska}
	\end{subfigure}
	\begin{subfigure}{0.5\textwidth}
        \vskip\baselineskip
        \begin{minipage}[c]{0.49\textwidth}   
            \centering 
			\includegraphics[width=\textwidth]{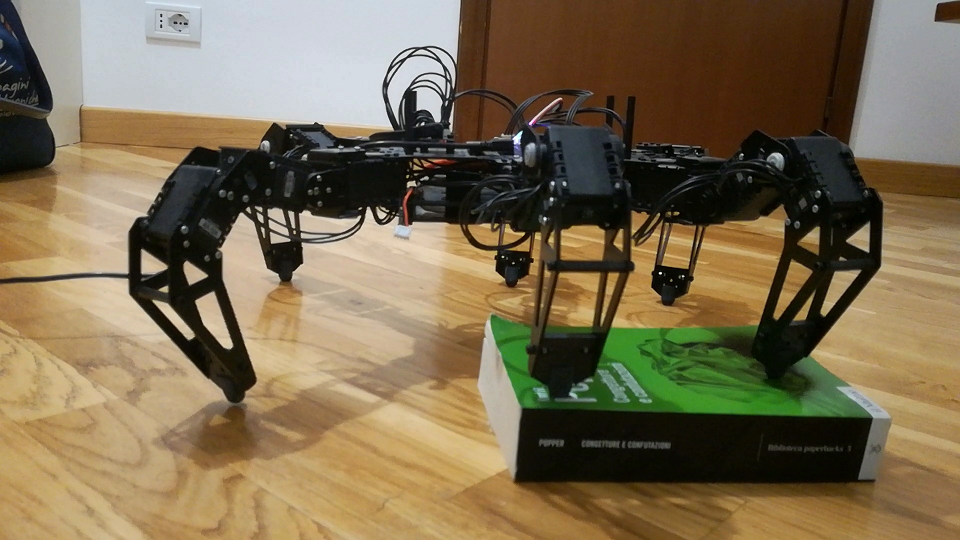}
        \end{minipage}
        \hfill
        \begin{minipage}[c]{0.49\textwidth}   
			\centering 
			\includegraphics[width=\textwidth]{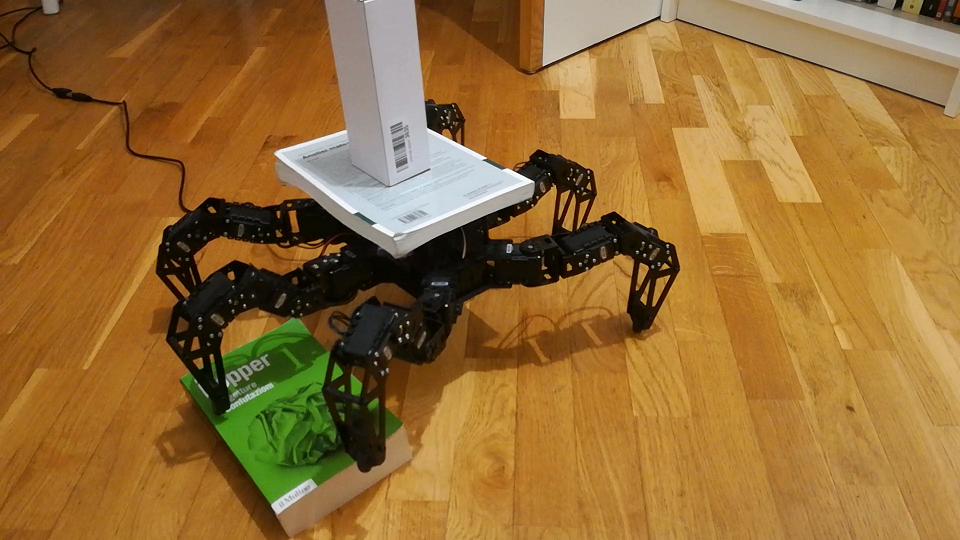}
		\end{minipage}
		\caption{Feed-back control logic aided walking}
		\label{obstacletaskb}
	\end{subfigure}
\caption{Physical applications of obstacle recognition capability. On top are shown performances of NUKE controller, on bottom the closed-loop control logic is used instead.}
\end{figure}

As shown in Figure \ref{obstacletaska} while the stock controller puts the robot in an instability stride, the novel closed loop solution is indeed able to maintain the body horizontal and by not overstepping onto the solid object all legs hold to the ground assuring equilibrium (Figure \ref{obstacletaskb}). These aspects are further highlighted when a payload is mounted on top of it, showing how stability of movement transfers to stability of robot body. Robustness of walking stride made in fact possible to effectively disregard completely the presence of the obstacle in the robot path.

\begin{figure}[t]
	\begin{minipage}[c]{0.4\textwidth}
	   \centering
	   \includegraphics[width=1\textwidth]{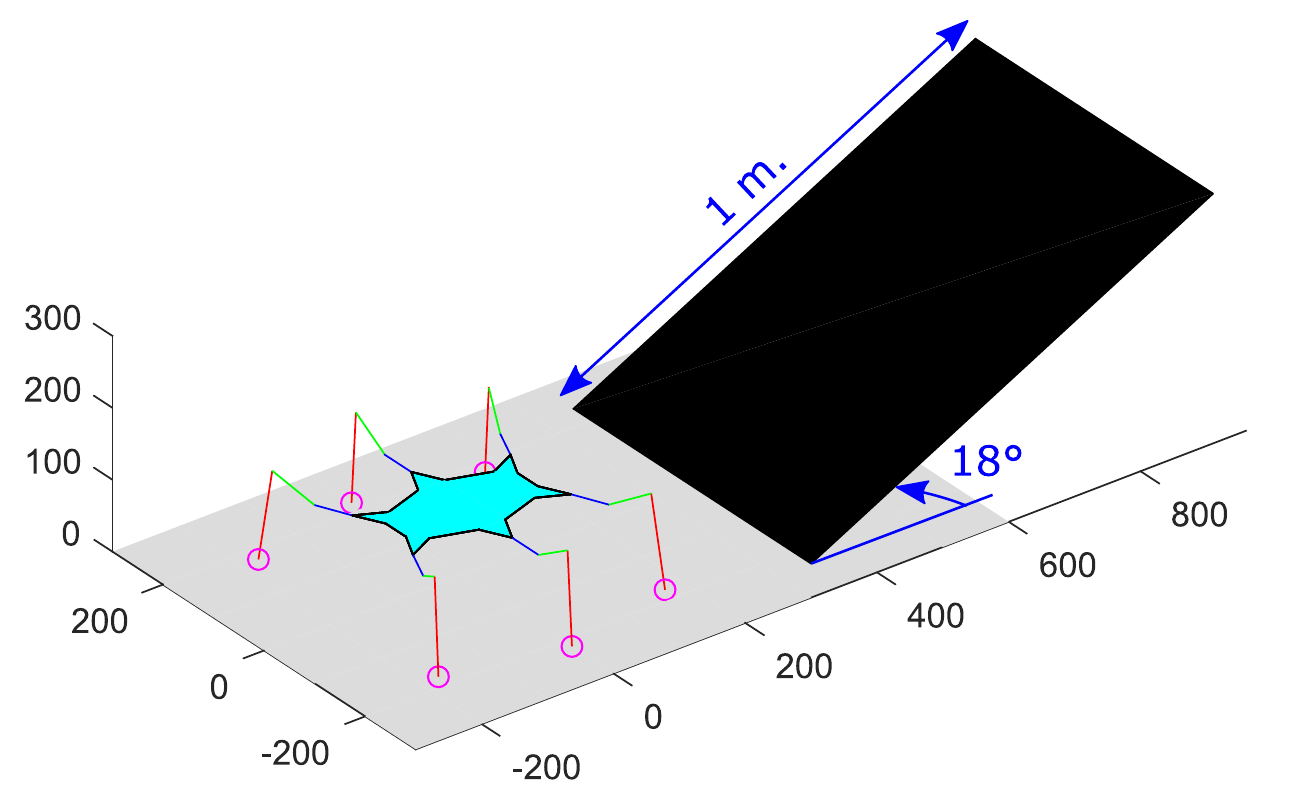}
	\end{minipage}
 \vskip\baselineskip
	\begin{minipage}[c]{0.4\textwidth}
	   \centering
	   \includegraphics[width=1\textwidth]{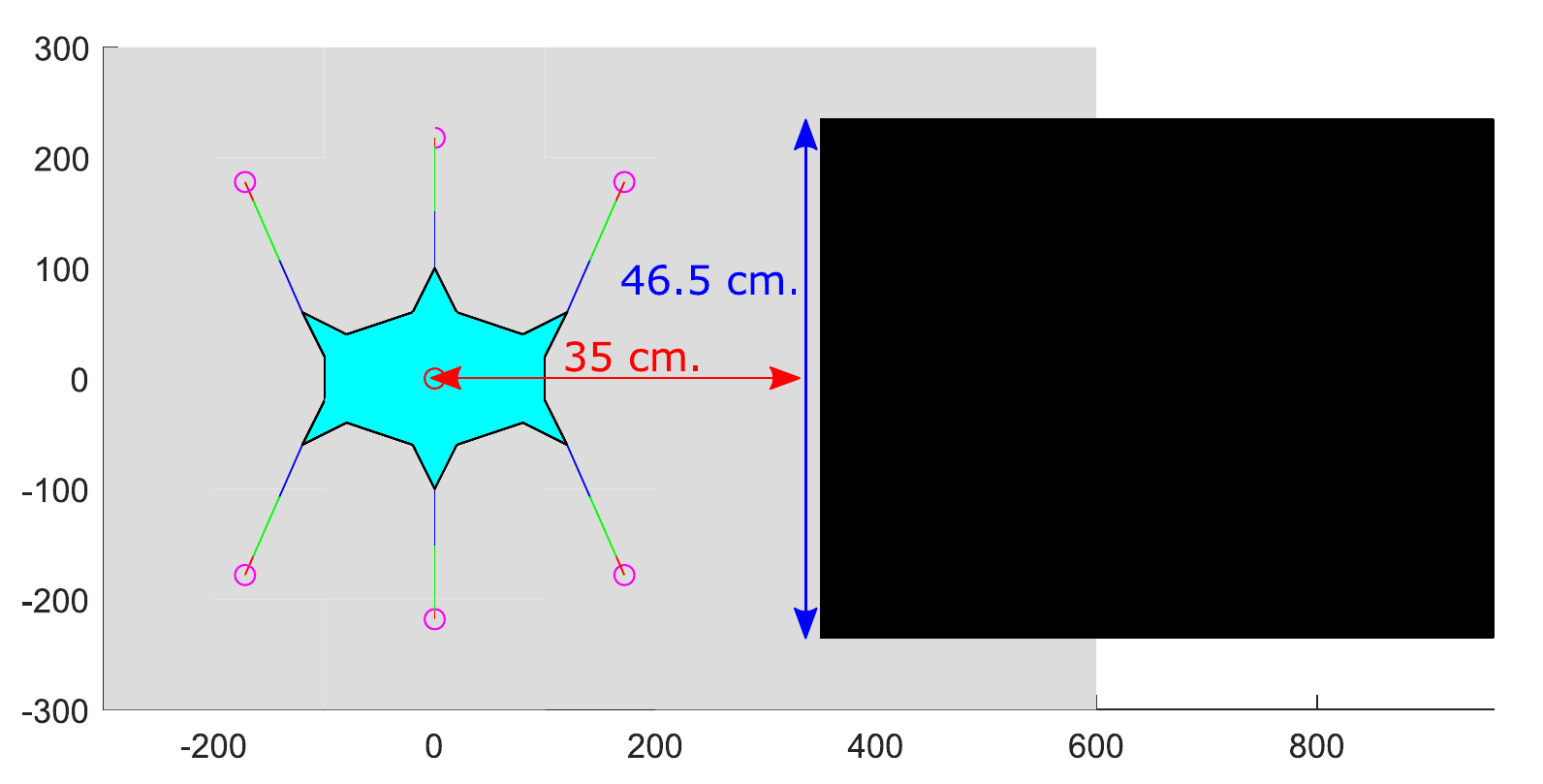}
	\end{minipage}
\caption{The ramp climbing task setup dimensions, angulation and position}
\label{rampscheme}
\end{figure}

\begin{figure}[t]
\begin{center}
\includegraphics[width=0.5\textwidth]{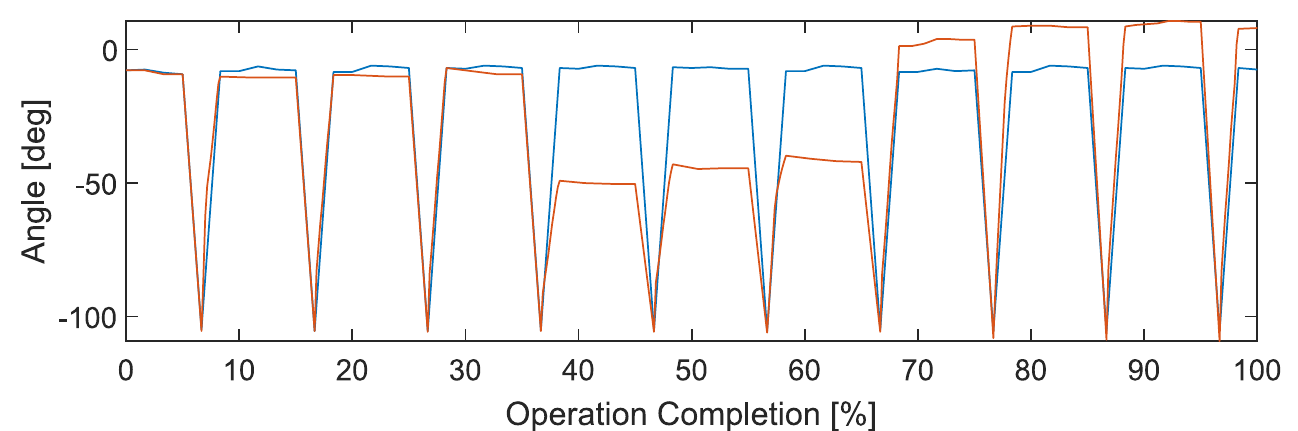}
\caption{Right-middle femur actuator angle comparison between operations. In blue color the operation employs the NUKE controller, in red color the feed-back control algorithm is used instead.\label{femur16loadcomp}}
\end{center}
\end{figure}



A comparison of the front-right and middle-right femur servo angle variation through the entire operation between NUKE and feed-back control logics is shown in Figure \ref{femur16loadcomp}. It is evident how thanks to the feed-back control algorithm, in the middle of the stride cycle the obstacle was found by the legs and that resulted into the respective femur actuator to stop on its tracks. \\
In Figure \ref{obstacletaska} it is also possible to check the `blindness' of the robot while being moved by the stock controller: Despite being in a completely unstable position, the angular feedback of the servomotors still reports no extra-ordinary values, and therefore the robot still attempts to walk as if no obstacle were in its path in the first place.

\subsection{Ramp Climbing}

In the ramp climbing task, the robot is positioned right in front of the angular interface, with the instruction of a forward constant movement. The setup scheme is presented in Figure \ref{rampscheme}. In order to request the robot body to remain horizontal the $Q_{ter}$ matrix is built as \eqref{Qternew} rather than \eqref{Qterdef}.

\begin{equation}
\label{Qternew}
Q_{ter,i} = D_{ter,i}
\end{equation}

\begin{figure}[t]
	\begin{subfigure}{0.5\textwidth}
		\centering
		\begin{minipage}[c]{0.49\textwidth}
			\centering
			\includegraphics[width=\textwidth]{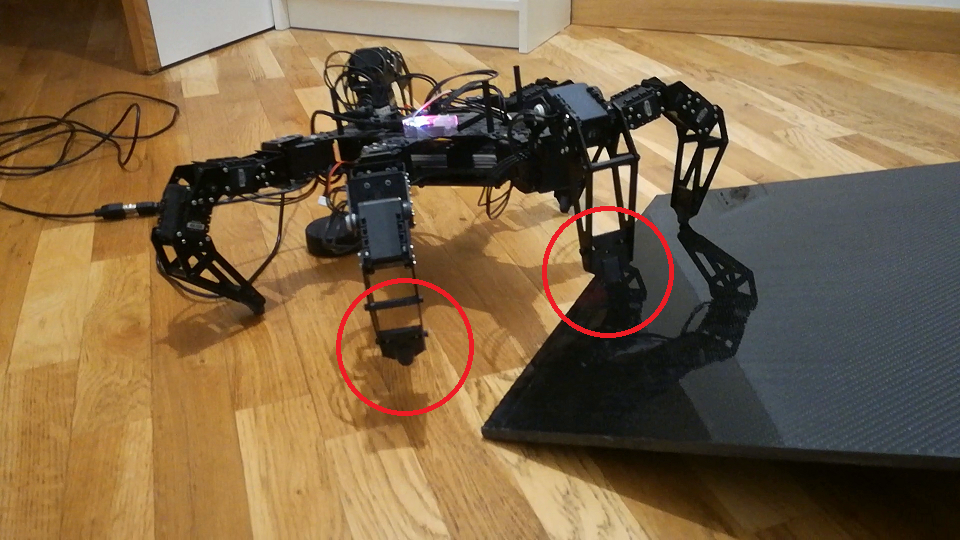}
		\end{minipage}
	\hfill
		\begin{minipage}[c]{0.49\textwidth}  
			\centering 
			\includegraphics[width=\textwidth]{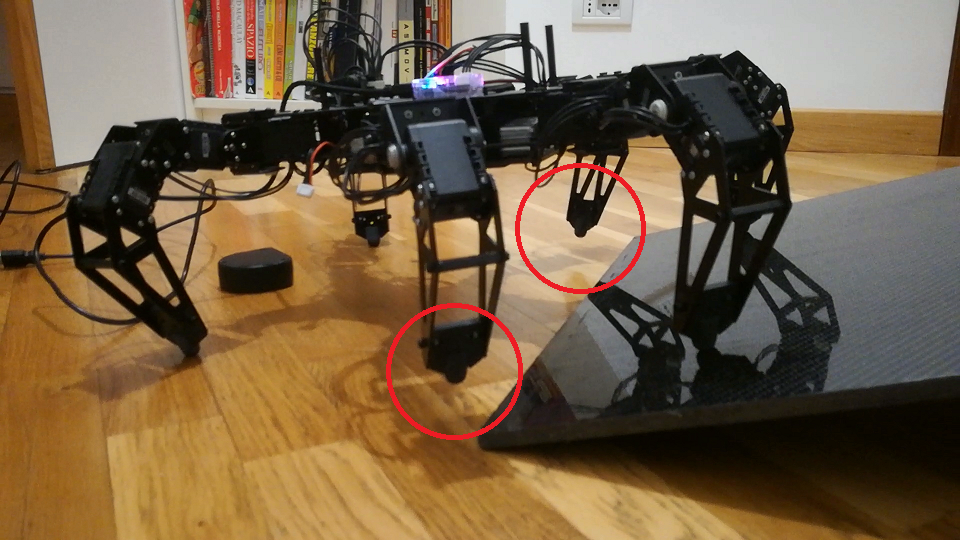}
		\end{minipage}
		\caption{Feed-forward only controller stability problems}
		\label{rampclimbtaska}
	\end{subfigure}
	\begin{subfigure}{0.5\textwidth}
        \vskip\baselineskip
        \begin{minipage}[c]{0.49\textwidth}   
            \centering 
			\includegraphics[width=\textwidth]{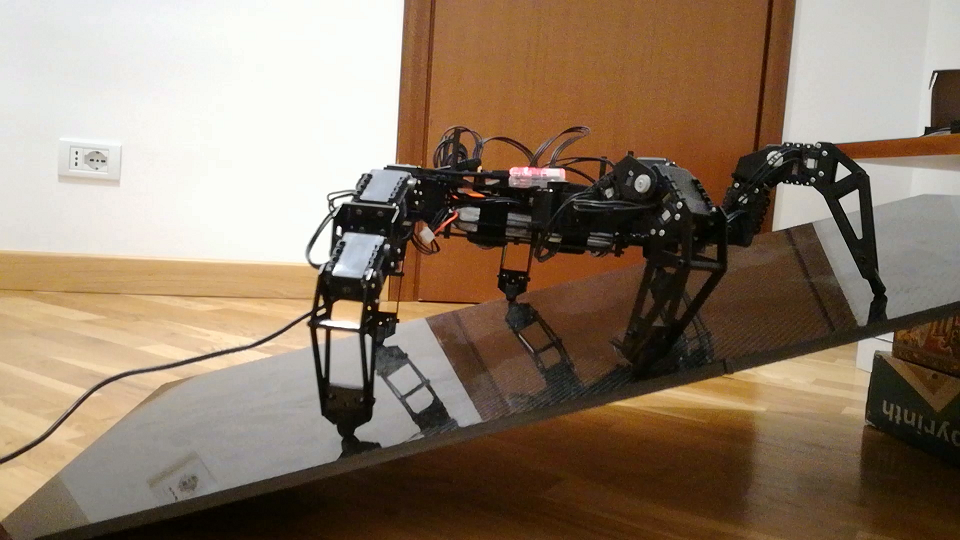}
        \end{minipage}
        \hfill
        \begin{minipage}[c]{0.49\textwidth}   
			\centering 
			\includegraphics[width=\textwidth]{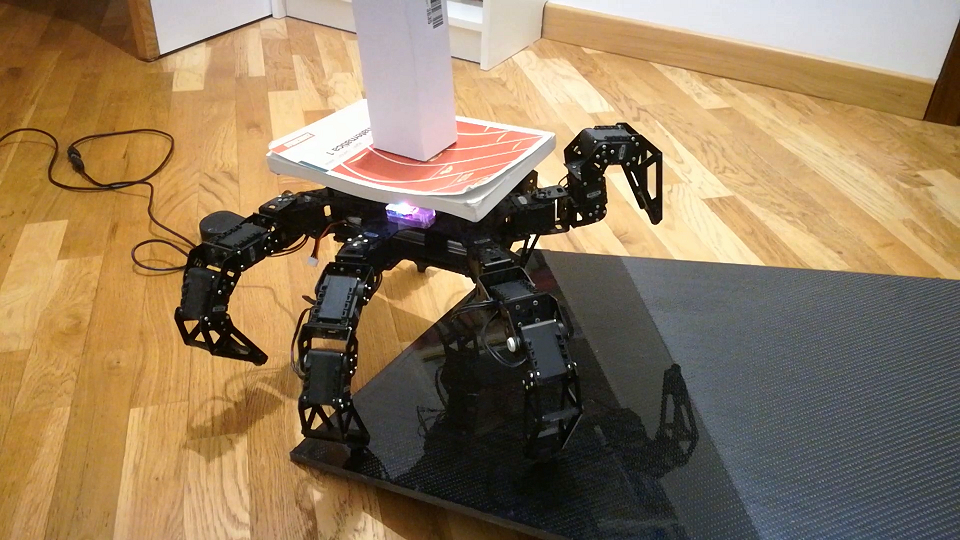}
		\end{minipage}
		\caption{Feed-back control logic performances}
		\label{rampclimbtaskb}
	\end{subfigure}
\caption{Physical applications of the closed loop control architecture. On top the feed-back algorithm is absent, on bottom the full control logic is used instead.}
\end{figure}

The most challenging part of the task is the traversal of the interface between flat plane and ramp, as in that location desync effects are present the most. \\
This is clearly visible in Figure \ref{rampclimbtaska}, where when the control feedback is absent then legs are left hanging quite often, leading to movement unsteadiness and ultimately increased traversal difficulty. By inducing the novel control architecture, as shown in Figure \ref{rampclimbtaskb}, the robot is instead able to climb the ramp correctly after successfully traversing the interface while maintaining an horizontal body pose. The task is done with such gait steadiness that even when carrying a payload the robot is able to assure its stability.

\section{Conclusions}

This paper presents a simple but effective kinematic model through the manipulation of hexapod legs endpoints; at any time it is possible to access legs position in space and robot pose through the use of transformation and pose matrices. \\
An autonomous terrain-adapting algorithm is developed, able to automatically tune the robot body pose in such a way to assure its isolation from rough terrain asperities no matter the terrain type. \\
A full dynamic model comprising of ground interaction modelling is presented, with the possibility of a real time implementation while providing accurate estimation of servomotor torques. \\
Finally a terrain sensing algorithm is presented, able to correct instability situations coming from hardware's non-idealistic behaviour, as well as guaranteeing leg-ground reach based on a feedback architecture using estimated torque from the model and torque provided by servomotors as variables. \\
Novel control architecture, composed by terrain adapting, torque estimation and terrain sensing algorithms, is evaluated in term of applicability and performances by means of experimental tests conducted on PhantomX AX Metal Hexapod Mark II robotic platform and results are reported.


\bibliography{bibliography}

\end{document}